%% file: main.tex
\documentclass[10pt,journal,compsoc]{IEEEtran}

\usepackage{graphicx}
\usepackage{amsmath}
\usepackage{amssymb}
\usepackage{dsfont}
\usepackage{booktabs}
\usepackage{multirow}
\usepackage{makecell} 
\usepackage{colortbl}
\usepackage{threeparttable}
\usepackage{pifont}
\newcommand{\cmark}{\ding{51}}
\newcommand{\xmark}{\ding{55}}
\usepackage[misc]{ifsym}

\newcommand{\etal}{\textit{et al}. }
\newcommand{\ie}{\textit{i}.\textit{e}., }

\ifCLASSOPTIONcompsoc
  \usepackage[nocompress]{cite}
\else
  \usepackage{cite}
\fi

\ifCLASSOPTIONcompsoc
 \usepackage[caption=false,font=footnotesize,labelfont=sf,textfont=sf]{subfig}
\else
 \usepackage[caption=false,font=footnotesize]{subfig}
\fi

\usepackage{stfloats}

\usepackage[capitalize]{cleveref}
\crefname{section}{Sec.}{Secs.}
\Crefname{section}{Section}{Sections}
\Crefname{table}{Table}{Tables}
\crefname{table}{Tab.}{Tabs.}

\input{mytables.tex}

\begin{document}

\title{STMixer: A One-Stage Sparse Action Detector}

\author{Tao~Wu,
        Mengqi~Cao,
        Ziteng~Gao,
        Gangshan~Wu~\IEEEmembership{Member,~IEEE},
        Limin~Wang~\IEEEmembership{Member,~IEEE}
\IEEEcompsocitemizethanks{
\IEEEcompsocthanksitem T.Wu, M.Cao, Z.Gao, G.Wu and L.Wang are with the State Key Laboratory for Novel Software Technology, Nanjing University, Nanjing 210023, China. (E-mail: \{wt,mg20370004\}@smail.nju.edu.cn, gzt@outlook.com, \{gswu,lmwang\}@nju.edu.cn (Corresponding author: Limin Wang.)) 
\IEEEcompsocthanksitem L. Wang is also with the Shanghai Artificial Intelligence Laboratory, Shanghai 200232, China.
} }


\IEEEtitleabstractindextext{
\begin{abstract}
    Traditional video action detectors typically adopt the two-stage pipeline, where a person detector is first employed to generate actor boxes and then 3D RoIAlign is used to extract actor-specific features for classification. This detection paradigm requires multi-stage training and inference, and the feature sampling is constrained inside the box, failing to effectively leverage richer context information outside. Recently, a few query-based action detectors have been proposed to predict action instances in an end-to-end manner. However, they still lack adaptability in feature sampling and decoding, thus suffering from the issues of inferior performance or slower convergence. In this paper, we propose two core designs for a more flexible one-stage sparse action detector. First, we present a query-based adaptive feature sampling module, which endows the detector with the flexibility of mining a group of discriminative features from the entire spatio-temporal domain. Second, we devise a decoupled feature mixing module, which dynamically attends to and mixes video features along the spatial and temporal dimensions respectively for better feature decoding. Based on these designs, we instantiate two detection pipelines, that is, STMixer-K for keyframe action detection and STMixer-T for action tubelet detection. Without bells and whistles, our STMixer detectors obtain state-of-the-art results on five challenging spatio-temporal action detection benchmarks for keyframe action detection or action tube detection.
\end{abstract}

}

\maketitle

\IEEEdisplaynontitleabstractindextext

\IEEEpeerreviewmaketitle

\ifCLASSOPTIONcompsoc
\IEEEraisesectionheading{\section{Introduction}\label{sec:introduction}}
\else
\section{Introduction}
\label{sec:introduction}
\fi

\IEEEPARstart{V}{ideo} action detection~\cite{fat,hcstal,learningtrack,hfcn,multitwo,stagn} is a fundamental task in video understanding, which aims to recognize all action instances present in a video and localize them in both space and time. There are two different forms of this task: The first is keyframe action detection~\cite{ava,avakinetics}, which only requires the detection of action boxes on each individual keyframe; the second is action tube detection~\cite{ucf101,jhmdb,multisports}, which requires dense detection on all video frames and linking the boxes of the same person and action category across adjacent frames into an action tube.

Video action detection has drawn significant research attention due to its wide applications in areas like security monitoring and sports video analysis. Benefiting from the proposal of large-scale action detection benchmarks~\cite{ava,avakinetics,multisports} and the advances of video representation learning such as video convolution neural networks~\cite{c3d,s3d,i3d,r2p1d,slowfast,non-local,csn,v4d,x3d} and video transformers~\cite{mvit,mvitv2,vivit,videoswin,vidtr,bevt,vmae,VideoMAEv2}, action detection has made remarkable progress in recent years. However, there are still several critical issues with the current methods.

\label{para:two_basics}
Most current action detectors~\cite{acrn,vat,slowfast,aia,acarn} adopt the two-stage Faster R-CNN-alike detection paradigm~\cite{fasterrcnn}. First, an offline actor detector or a separate actor localization head is used to generate actor bounding boxes in advance. Second, the RoIAlign~\cite{maskrcnn} operation is applied on video feature maps to extract actor-specific features for action classification. This two-stage action detection pipeline requires large computing resources and exhibits relatively lower efficiency. Furthermore, the RoIAlign~\cite{maskrcnn} operation constrains the video feature sampling inside the actor bounding box and lacks the flexibility of capturing context information in its surroundings. To enhance the RoI features, recent works resort to extra modules that aggregate context information from other actors and detected objects~\cite{aia} or global feature maps~\cite{acarn}, which further increases the computational burden.

\begin{figure}[t]
    \centering
    \includegraphics[width=0.45\textwidth]{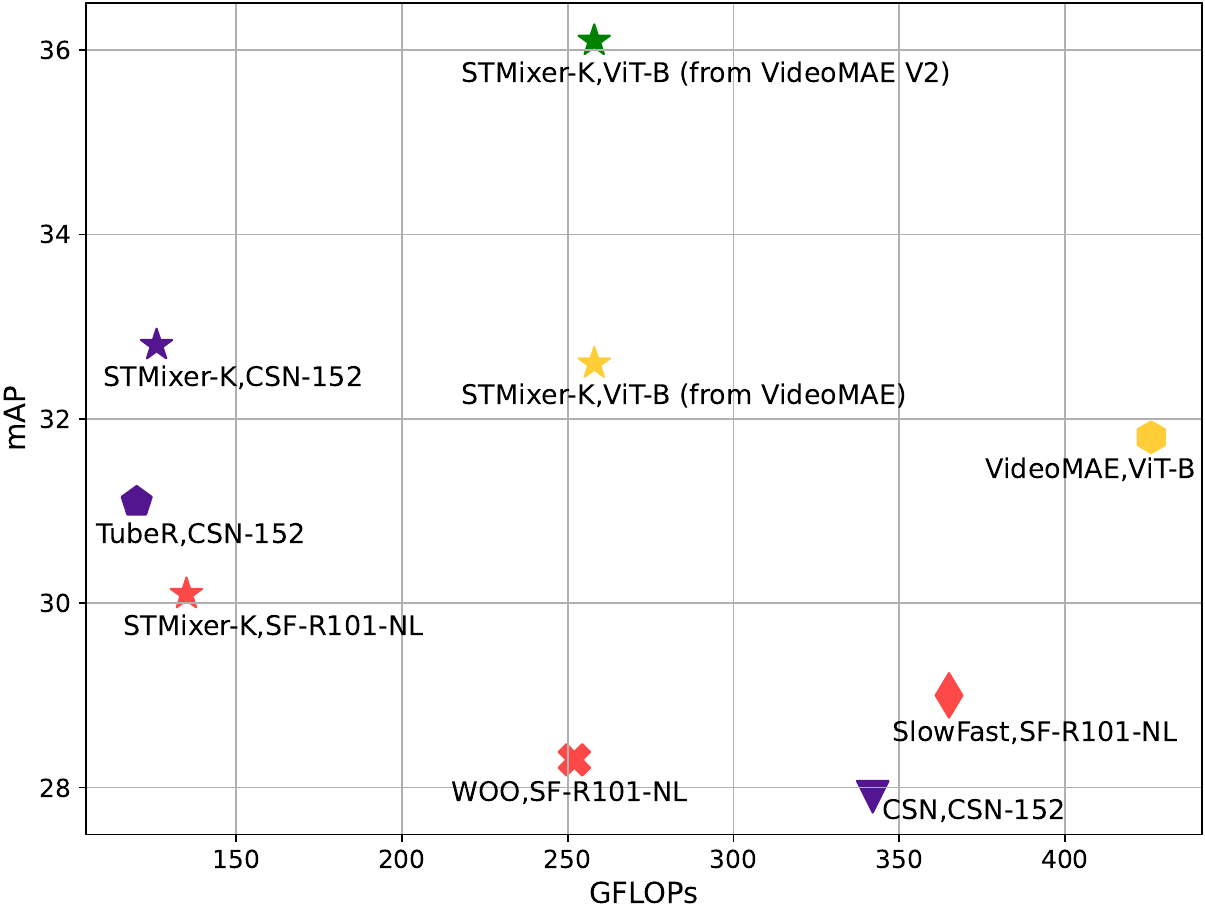}
    \caption{Comparison of mAP versus GFLOPs between different keyframe action detectors on AVA v2.2. The GFLOPs of CSN, SlowFast, and VideoMAE are the sum of Faster RCNN-R101-FPN detector GFLOPs and classifier GFLOPs. Different methods are marked by different makers and models with the same backbone are marked in the same color. The results of CSN are from \cite{TubeR}. Our STMixer-K achieves the best effectiveness and efficiency balance.}
    \label{fig:flops}
\end{figure}

Recently, several query-based action detectors~\cite{woo,TubeR} have been proposed to predict action instances in an end-to-end fashion. These detectors represent action proposals as a small set of learnable queries and formulate action detection as a set prediction task. WOO~\cite{woo} utilizes RoIAlign~\cite{maskrcnn} for action feature sampling and employs dynamic convolution to enhance query-specific features. However, constrained feature sampling results in inadequate contextual information. Building on the DETR~\cite{DETR} framework, TubeR~\cite{TubeR} applies global attention~\cite{attention} to the entire feature map for feature gathering. Nevertheless, the computational expense of global attention hinders its utilization of multi-scale features. For feature decoding, TubeR adopts a feed-forward network with static parameters. The features sampled by different queries are decoded in the same way, which lacks flexibility. In this paper, we argue that augmenting adaptability in feature sampling and decoding is imperative for enhancing query representation and facilitating the cast from query to action instance.

\begin{figure}[t]
    \centering
    \includegraphics[width=\linewidth]{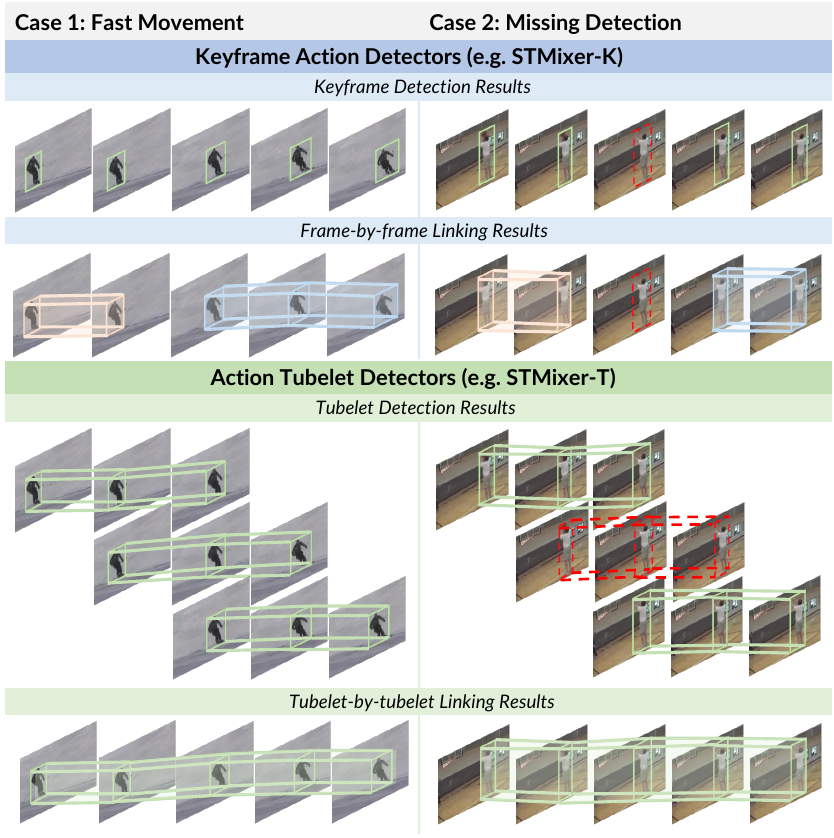}
    \caption{Comparisons between keyframe action detectors and action tubelet detectors in generating video-level action tubes. We show two failure cases of the keyframe action detectors. On the left, the person moves so fast that the two boxes in frames 2 and 3 have a small IoU, which causes the action instance to be mistakenly split into 2 tubes. On the right, a missed detection in frame 3 causes a linking failure. However, linking tubelets produces correct action tubes in both cases.}
    \label{fig:link}
\end{figure}

In addition, most recent action detectors~\cite{acarn,aia,woo} only pursue keyframe detection results and ignore the requirements of video-level action tube detection. However, action tube detection is also an important branch of action detection. Although it is feasible to link keyframe results into action tubes by linking every two temporally adjacent boxes with the same action category and high spatial IoU, as shown in Fig.~\ref{fig:link}, such linking is unreliable and easy to fail for two reasons: 1) The boxes on two adjacent frames in an action tube do not necessarily have a large IoU. For example, when a person moves very fast, the two boxes may overlap very little, which may lead to a failure in linking and cause an action instance to be mistakenly split into multiple tubes. In some extreme cases such as when two people exchange positions, it may even cause two boxes belonging to different instances to be linked together. 2) It is not robust and a missed detection or classification error in any single frame can cause linking failure. For more reliable action tube linking, action tubelet detectors are preferred. Action tubelet detectors predict tracked action boxes on each video clip of $T$ consecutive frames. Two adjacent clips have $T-1$ frames overlapping. We can link tubelets into action tubes by linking every two tubelets with the same action category and high spatial IoU on overlapped frames. This linking strategy can effectively avoid the above two problems. Two tubelets of the same action tube must have a high IoU on overlapped frames because they point to the same actor at the temporal point. A missed detection or classification error on a certain video clip does not necessarily cause linking failure, because the next video clips could still have frames overlapping.

Based on the above observation and analysis of existing action detectors, we present a new query-based one-stage video action detection framework, coined STMixer. Inspired by AdaMixer~\cite{AdaMixer} for image object detection, we propose to sample and decode features from the complete video spatio-temporal domain in a more flexible manner. Specifically, we propose two core designs. First, we present a query-guided adaptive feature sampling module to mine a set of discriminative features from the entire spatio-temporal domain, which captures tailored context for each specific query. Second, we devise a spatio-temporal decoupled feature mixing module to extract discriminative representation from sampled features. It is composed of an adaptive spatial mixer and an adaptive temporal mixer in parallel to focus on appearance and motion information, respectively. The mixing parameters are all dynamically generated based on queries, utilizing the prior information contained in the query to guide the weights attached to different feature channels and points. Coupling these two designs with a video backbone yields a simple, neat, and effective end-to-end action detection framework.

Based on the STMixer framework, we first instantiate the STMixer-K pipeline for keyframe action detection. We evaluated the STMixer-K detector on AVA~\cite{ava} and AVA-Kinetics~\cite{avakinetics} benchmarks which focus on keyframe action detection. AVA and AVA-Kinetics have many action categories related to actor-actor or actor-object interaction, which requires high context and interaction modeling capabilities of the detector. Benefiting from flexible feature sampling and decoding mechanism, our STMixer-K can see out of the actor box, better mining and decoding discriminative features from the entire spatio-temporal domain to model the context and interaction. Our STMixer-K achieves state-of-the-art performance on AVA and AVA-Kinetics benchmarks with remarkable improvements compared with former methods. As shown in Fig.~\ref{fig:flops}, our STMixer-K also achieves best effectiveness and efficiency balance.

Beyond keyframe action detection, our STMixer framework can also be easily extended into an action tubelet detector, coined STMixer-T,  to meet the requirement of video-level action tube detection. There are two main challenges in designing a tubelet detector: 1) Tubelet detection requires predicting action boxes densely, so the inputs are usually composed of continuous video frames, covering a short temporal range.  It is challenging to leverage temporal information effectively in a short video clip with limited temporal motion and variance. 2) A tubelet needs to track the same person in different frames accurately, and at the same time, the predicted action box of each frame needs to be adaptive to the position and posture of the person in the current frame. For better temporal modeling, our proposed temporal feature mixing can better capture subtle changes in each frame and model motion patterns. For accurate person tracking and box prediction, we propose to use a set of queries with the same initialization to independently perform feature sampling and spatial mixing in the feature space corresponding to each frame. Initially using the same queries in different frames can make it easier for them to focus on the same person. Independent feature sampling and spatial mixing in each frame make it easier to adjust the predicted action box according to frame-specific information. We evaluate our tubelet detector STMixer-T on three popular video action detection benchmarks focusing on action tube detection, namely UCF101-24~\cite{ucf101}, JHMDB51-21~\cite{jhmdb}, MultiSports~\cite{multisports}. Our STMixer-T achieves leading results on UCF101-24, JHMDB51-21 and MultiSports benchmarks, surpassing the previous methods by a large margin.

In summary, we make four major contributions:
\begin{itemize}
    \item We present a new one-stage sparse action detection framework in videos (STMixer). Our STMixer is easy to train in an end-to-end manner and efficient to deploy for action detection in a single stage without the dependency on an extra actor detector.
    \item We devise two flexible designs to yield a powerful and efficient action detection framework. The adaptive sampling can select the discriminative feature points and the adaptive feature mixing can enhance spatio-temporal representations.
    \item Based on the STMixer framework, we instantiate the STMixer-K detector for keyframe action detection and the STMixer-T detector for action tube detection. For the latter, we propose to use queries with the same initialization to perform feature sampling and spatial mixing in different frame feature spaces for better actor tracking and box prediction.
    \item Our STMixer detectors achieve a new state-of-the-art performance on five challenging action detection benchmarks, in both keyframe action detection and action tube detection. 
\end{itemize}

A preliminary version~\cite{STMixer} of this work has been accepted to CVPR 2023. In this journal version, we extend our previous work in a number of aspects: 1) Beyond the keyframe action detector STMixer-K in the previous work, we instantiate a new detection pipeline STMixer-T to produce dense tubelet detection results, which benefits the linking of video-level action tubes in accuracy and robustness. 2) To meet the challenges of tubelet detection, we make adaptive improvements to query definition and ways of feature sampling and mixing in STMixer-T. Specifically, for accurate actor localization and tracking in each tubelet, we propose to use a series of queries with the same initialization to guide the feature sampling and spatial mixing process in each corresponding separate frame feature space. For more effective temporal modeling, a parallel temporal mixing branch is adopted to capture subtle changes between consecutive frames and motion patterns. 3) We conduct extensive ablation experiments to investigate the influence of our proposed feature sampling and mixing mechanism on both STMixer-K and STMixer-T. 4) Besides leading performance on the AVA and AVA-Kinetics datasets achieved by STMixer-K, in this version, we validate our proposed tubelet detector STMixer-T on UCF101-24, JHMDB51-21, and MultiSports for action tube detection and set new state-of-the-art records on these benchmarks. Qualitative analysis is also provided to demonstrate the advantage of STMixer-T in actor localization and tracking and action tube boundary judgment.

\section{Related Work}
\label{sec:relatedwork}

\subsection{Video Action Detection}

\noindent \bf Keyframe Action Detectors. \rm Keyframe action detection~\cite{ava,avakinetics} requires predicting actor boxes and corresponding action categories on each keyframe of a video. These detectors usually take a video clip centered at the keyframe as input for more spatio-temporal context. AVA~\cite{ava} is a benchmark focusing on keyframe action detection with frame-level action instances annotated on keyframes at 1FPS.

Currently, the most performant action detectors~\cite{lfb,slowfast,mvit,aia,acarn,hit,vmae,memvit,carcnn} on the AVA dataset rely on an auxiliary actor detector to perform actor localization on the keyframes. Typically, the powerful Faster RCNN-R101-FPN~\cite{fasterrcnn} detector is used as the human detector, which is first pre-trained on the COCO~\cite{coco} dataset and then fine-tuned on the target AVA~\cite{ava} dataset. With actor bounding boxes predicted in advance, the action detection problem is reduced to a pure action classification problem. The RoIAlign~\cite{maskrcnn} operation is applied on the 3D feature maps extracted by a video backbone to generate actor-specific features. SlowFast~\cite{slowfast} and MViT~\cite{mvit} directly use the RoI features for action classification. However, RoI features only contain the information inside the bounding box but overlook context and interaction information outside the box. To remedy this inherent flaw of RoI features, recent works resort to extra modules that introduce more context information, which brings extra computational costs. AIA~\cite{aia} uses a series of interaction blocks to aggregate context information from other actors and additional detected objects in the scene. ACARN~\cite{acarn} employs relation reasoning modules to attach context information extracted from the global feature maps to each actor's ROI feature. Though showing leading detection performance on the AVA benchmark, the models with an extra human detector require an additional training process and training data for actor detection. The training overhead is large and the two-stage inference is less efficient. Besides, they suffer from the aforementioned issue of fixed RoI feature sampling.

Methods of another research line use a single model to perform action detection. Most of them~\cite{yowo,acrn,vat,woo} still follow the two-stage pipeline but simplify the training process by jointly training the actor proposal network and action classification network in an end-to-end manner. Among them, WOO~\cite{woo} is a query-based method that adopts two separate query-based heads for actor localization and action classification respectively. These methods still have the issue of fixed RoI feature sampling. As a supplement, VTr~\cite{vat} attends RoI features to full feature maps while ACRN~\cite{acrn} introduces an actor-centric relation network for interaction modeling.

In this paper, we propose a new query-based keyframe action detector STMixer-K. STMixer-K does not rely on additional human detectors and predicts action instances end-to-end in one stage. STMixer-K adaptively samples discriminative features from a multi-scale spatio-temporal feature space and decodes them with a more flexible scheme under the guidance of queries for better actor localization and action classification on the keyframes.

\noindent \bf Action Tubelet Detectors. \rm Action tube detection~\cite{ucf101,jhmdb,multisports} requires dense detection on all video frames and linking the boxes of the same person and action category across adjacent frames into an action tube. Though it is possible to link keyframe detections into action tubes, the linking results can be unreliable. For more reliable action tube generation, tubelet detectors are preferred. 

Earlier tubelet detectors~\cite{act,rtprn,2in1,tacnet,t-cnn} are adapted from anchor-based object detectors~\cite{ssd,yolo,focalloss,fastrcnn,fasterrcnn,RichFeature,spp}. They introduce the idea of anchors into tubelet detection by extending the 2D anchor boxes into 3D anchor cuboids. Following the two-stage Faster R-CNN~\cite{fasterrcnn} pipeline, T-CNN~\cite{t-cnn} uses a proposal network to generate proposal boxes based on anchors and then replicate them along the temporal dimension to form proposal cuboids which are then refined with ROI features. These anchors are generated by clustering on each specific dataset. ACT~\cite{act} regresses anchor cuboids into tubelets based on the features of each frame and then uses stacked frame features to predict the action category. STEP~\cite{step} takes a progressive approach, refines proposal cuboids through multiple steps, and gradually increases their length to obtain richer temporal information. These methods are sensitive to pre-set anchors, which require careful tuning for each dataset. They make a strong assumption that the position and posture of actors will not change greatly across frames, which limits their ability to cope with fast and complex actions. 

MOC~\cite{moc} is built upon the anchor-free object detector CenterNet~\cite{centernet}. It regards each action tubelet as a trajectory of moving points. An image backbone is employed for frame feature extraction and the frame features are concatenated along the temporal axis to form the video feature maps. Each point on the feature maps is regarded as an action instance proposal. The bounding box and action scores of each point are predicted by convolution. MOC relies more on appearance features and lacks temporal and interaction modeling. Building on DETR~\cite{DETR} framework, TubeR~\cite{TubeR} is a query-based tubelet detector. TubeR applies global attention to single-scale video feature maps, neglecting multi-scale information which is important for detection tasks. The gathered features are transformed with static parameters, failing to decode the features with query-specific information. For box prediction in each frame, the features are sampled from the same spatial-temporal domain, which makes it more difficult for the queries to focus on the frame-specific features. This further exacerbates the problem of slow convergence and also affects the accuracy of the predicted boxes.

We instantiate an action tubelet detector named STMixer-T to deal with video action tube detection. STMixer-T does not require hand-crafted anchor cuboids, nor does it make any assumptions about human movements and trajectories. STMixer-T uses a set of queries with the same initialization to independently perform spatial feature sampling in the feature space of each frame to better adapt to the position and posture changes of the actors in each frame. A parallel temporal feature mixing branch is adopted for better temporal modeling.

\begin{figure*}[ht]
    \centering
    \includegraphics[width=17.5cm]{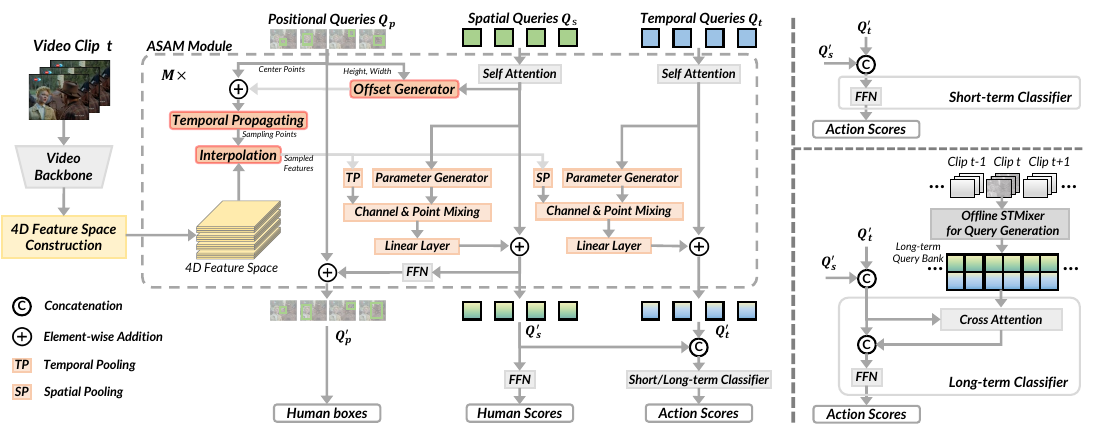}
    \caption{Pipeline of STMixer-K for keyframe action detection. On the left is the overall architecture of STMixer-K. A 4D feature space is constructed on the feature maps of the input video clip. The action decoder contains $M$ stacked ASAM modules. In each module, adaptive feature sampling is performed first. Specifically, a group of offsets is generated on each spatial query by a linear layer and added to the center point of the corresponding positional query, thus yielding sampling points on the keyframe. The sampling points are then temporally propagated and used as the index for feature sampling from the 4D feature space. After feature sampling, adaptive feature mixing is performed. For spatial mixing, temporal pooling is applied to the sampled features. The mixing parameters are generated on each spatial query by a linear layer. Channel and point mixing are performed sequentially. Finally, the mixed feature is transformed by a linear layer and used to update the spatial query. Temporal mixing is performed symmetrically. Optionally, a short-term or long-term classifier can be used for action score prediction, whose detailed structures are illustrated on the right. The long-term classifier refers to the query bank produced by an offline STMixer-K for long-term information.}
    \label{fig:framework}
\end{figure*}

\subsection{Query-based Detection}

Query-based object detectors~\cite{DETR,deformableDETR,sparsercnn,AdaMixer,anchordetr,efficientdetr,condtionaldetr,dino} represent objects as a set of learnable queries. Then, object detection is formulated as a query-based set prediction task, which is trained based on the bipartite matching between the predictions and ground-truth objects. DETR~\cite{DETR} is the first query-based detector, which aims to eliminate the need for many hand-designed components in object detection. DETR introduces the transformer~\cite{attention} architecture to object detection. The learnable object queries in the transformer decoder iteratively attend to encoded image features and are then decoded as bounding boxes and classification scores. However, as analyzed by ~\cite{sparsercnn}, TubeR is not a pure sparse method since each query has to interact with dense features over full images, and the expensive global attention and static feature decoding make its convergence very slow. Due to the limited feature map resolution, DETR performs unsatisfying on small objects. New attention module designs and new query-based paradigms are proposed to deal with these issues. Deformable DETR~\cite{deformableDETR} tackles the problem of slow convergence by making its attention modules only attend to a small set of key points around a reference point. The removal of global attention also enables the use of multi-scale feature maps. Sparse R-CNN~\cite{sparsercnn} introduces learnable proposal boxes besides learnable object queries. It follows the paradigm of Cascade R-CNN~\cite{cascadercnn} to iteratively refine the proposal boxes with RoI features and proposes the dynamic instance interaction head to model the relation between global context and RoI features. Sparse RCNN~\cite{sparsercnn} and Deformable DETR~\cite{deformableDETR} are pure sparse detectors as their object queries only interact with a small set of sampled features.

Query-based object detectors have inspired research on many other detection tasks like human-object interaction detection~\cite{hoidis,hoitr,hoitrans,conhoi,lookhoi,mstrhoi,sahoitr,catranhoi,distillhoi}, temporal action detection~\cite{aftad,lstad,e2etad,stptad,oadtrtad,relaxtad}, and spatio-temporal action detection~\cite{woo,TubeR}. Our STMixer is also inspired by query-based object detectors. However, we propose new core designs for the video action detection task, including the construction of 4D video feature space, adaptive spatio-temporal feature sampling and decoding.

\section{Method}

\begin{figure}[t]
    \centering
    \includegraphics[width=9cm]{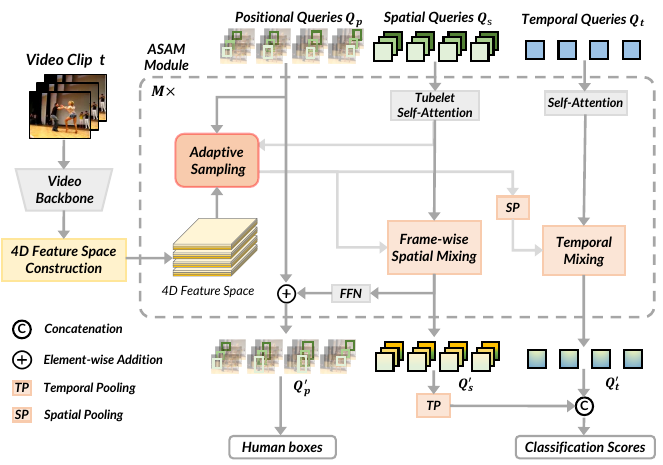}
    \caption{Pipeline of STMixer-T for action tubelet detection. A video clip composed of $T$ consecutive frames is input to the video backbone for feature extraction. A 4D feature space is constructed on the feature maps. We perform adaptive feature sampling and spatial mixing under the guidance of each spatial query in its corresponding frame feature space. A parallel temporal mixing branch is adopted for temporal modeling.}
    \label{fig:tubelet}
\end{figure}

This section presents our one-stage sparse action detection framework, STMixer. As shown in Fig.~\ref{fig:framework} and Fig.~\ref{fig:tubelet}, the STMixer framework comprises a video backbone for feature extraction, a feature space construction module, and a sparse action decoder composed of $M$ adaptive sampling and adaptive mixing (ASAM) modules followed by prediction heads. We first use the video backbone to extract feature maps for the input video clip and a 4D feature space is constructed based on the feature maps. Then, we develop the sparse action decoder with a set of learnable queries. Under the guidance of these queries, we perform adaptive feature sampling and feature mixing in the video feature space. The queries are updated iteratively. Finally, we decode each query as a detected action instance with prediction heads. We will describe the core components of the STMixer framework in detail and highlight the difference in implementation of the \textit{keyframe action detector} \textbf{STMixer-K} and \textit{action tubelet detector} \textbf{STMixer-T} in the next subsections.

\subsection{4D Feature Space Construction} \label{para:feature_space}

For \textit{keyframe action detection}, \textbf{STMixer-K} samples a video clip composed of $T$ frames centered at each keyframe. To cover a larger temporal range, strided sampling is performed. For \textit{action tubelet detection}, \textbf{STMixer-T} takes every video clip of $T$ consecutive frames as an input. Two adjacent clips have $T-1$ frames overlapping. The video clips are then input into a video backbone for feature extraction.

\begin{figure}[t]
    \centering
    \includegraphics[width=8cm]{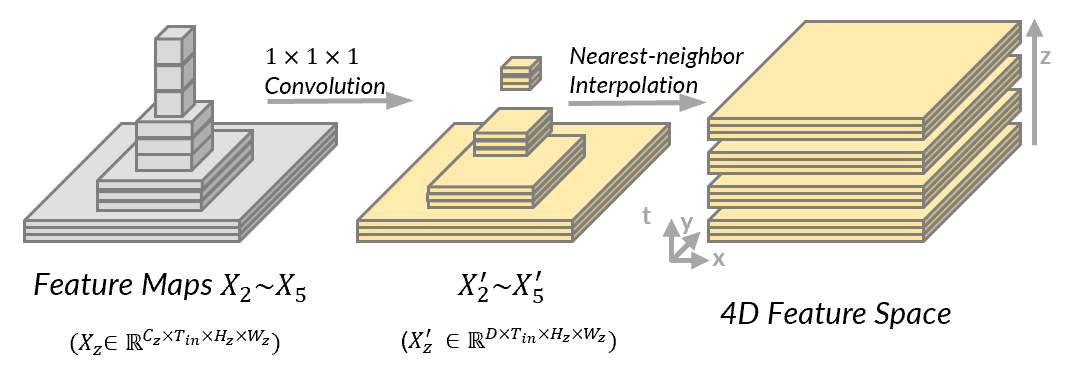}
    \caption{4D feature space construction for hierarchical video backbone. We construct 4D feature space on multi-scale 3D feature maps from hierarchical video backbone by simple lateral convolution and nearest-neighbor interpolation. The four dimensions of the 4D feature space are x-, y-, t-axis, and scale index z. }
    \label{fig:featurespace}
\end{figure}

\noindent \bf Hierarchical CNN backbone. \rm Formally, let $X_z \in \mathbb{R}^{C_z\times T_{in}\times H_z\times W_z}$ denote the feature map of convolution stage $z$ of the hierarchical backbone, where $z \in \{2,3,4,5\}$, $C_z$ stands for the channel number, $T_{in}$ for temporal length, $H_z$ and $W_z$ for the spatial height and width. Typically, $T_{in}<T$ as the video backbone performs temporal downsampling to aggregate temporal information. For \textit{tubelet detection}, we remove the temporal downsampling in the backbone to maintain the temporal length of the output feature maps, thus $T_{in}=T$ in this situation. The stage index $z$ can be seen as the scale index of the feature map as $X_z$ has the spatial downsampling rate of $2^z$. We first transform each feature map $X_z$ to the same channel $D$ by $1 \times 1\times 1$ convolution. Then, we rescale the spatial shape of each stage feature map to $H_2 \times W_2$ by simple nearest-neighbor interpolation and align them along the x- and y-axis. The four dimensions of the constructed feature space are x-, y-, t-axis, and scale index z, respectively. This process is illustrated in Fig.~\ref{fig:featurespace}.

\noindent \bf Plain ViT backbone. \rm To make STMixer framework compatible with plain ViT~\cite{ViT} backbone, inspired by ViTDet~\cite{VitDet}, we construct 4D feature space based on the feature maps from the last Transformer layer of the ViT backbone. Specifically, with the output feature maps of the default downsampling rate of $2^4$, we first produce hierarchical feature maps $\left\{X_{z}\right\}$ of the same channel number $D$ using convolutions of spatial strides $\left\{\frac{1}{4},\frac{1}{2},1,2 \right\}$, where a fractional stride indicates a deconvolution. Then we also rescale each feature map to the spatial size of $H_2 \times W_2$.

\subsection{Query Definition} \label{para:query_def}

We define the action queries in a disentangled fashion. Specifically, we factorize action queries into spatial queries $Q_s \in \mathbb{R}^{N\times L \times D}$, positional queries $Q_p \in \mathbb{R}^{N\times L\times 4}$ and temporal queries $Q_t \in \mathbb{R}^{N\times D}$. $N$ represents the number of queries while $D$ denotes the dimension of each query. $L$ denotes the temporal length of the output boxes. For \textit{keyframe action detection}, $L=1$ as it only requires predicting action boxes on the keyframe. For \textit{action tubelet detection}, $L=T$ as it requires predicting action boxes densely on each frame of the input clip.

We use spatial queries $Q_s$ and temporal queries $Q_t$ to refer to the spatial content and temporal content respectively. The positional queries $Q_p$ are for position prediction and reference. Each positional query $q{}_{p}$ is formulated as a proposal box vector $(x,y,z,r)$. Specifically, $x$ and $y$ stand for the x- and y-axis coordinates of the box center, and $z$ and $r$ denote the logarithm of its scale (\ie the area of the bounding box) and aspect ratio. $z$ can be used as the scale index. The positional queries $Q_p$ are initialized in such a way that every box vector is the whole video frame.

\subsection{Adaptive Spatio-temporal Feature Sampling} \label{para:adasample}

\noindent \bf Sampling points generation. \rm Different from previous work~\cite{slowfast,aia,woo,acarn} that samples RoI features by pre-computed proposal boxes, we sample actor-specific features adaptively from the aforementioned 4D feature space under the guidance of spatial queries. Specifically, given a spatial query $q_{s}$ and the corresponding positional query $q_{p}$, we regard the center point $(x,y,z)$ of the proposal box as the reference point, and regress $P_{in}$ groups of offsets along x-, y-axis and scale index z on query $q_{s}$, using a linear layer:
\begin{equation}
\begin{gathered}
    \left\{ \left( \bigtriangleup x_{i},\bigtriangleup y_{i},\bigtriangleup z_{i} \right)  \right\} = \text{Linear} (q_{s}),\\
    \text{where}\ i\in \mathbb{Z}\ \text{and}\ 1\leqslant i\leqslant P_{in},
\end{gathered}
\end{equation}
where $P_{in}$ is the number of sampling points of each query. Then, the offsets are added to the reference point, thus $P_{in}$ spatial feature points are obtained:
\begin{equation}
    \begin{cases}\widetilde{x}_{i}=x+\bigtriangleup x_{i}\cdot 2^{z-r},\\ 
    \widetilde{y}_{i}=y+\bigtriangleup y_{i}\cdot 2^{z+r},\\ 
    \widetilde{z}_{i}=z+\bigtriangleup z_{i}.\end{cases}  
\end{equation}
where $2^{z-r}$ and $2^{z+r}$ are the width and height of the box respectively. We offset the spatial position of sampling points with respect to the width and height of the box to reduce the learning difficulty.

For \textit{keyframe action detection}, we propagate sampling points along the temporal axis, thus obtaining $T_{in}\times P_{in}$ points to sample from the 4D feature space. In our implementation, we simply copy these spatial sampling points along the temporal dimension because the current keyframe action detection dataset (AVA) yields temporal slowness property that the motion speed of the actors is very slow.

For \textit{action tubelet detection}, each spatial query $q_{s}$ generates $P_{in}$ sampling points in the 3D feature space of its corresponding temporal frame, thus obtaining $T_{in}\times P_{in}$ points in total for a tubelet.

\noindent \bf Feature sampling. \rm Given $T_{in}\times P_{in}$ sampling points, we use them as the positional index to sample instance-specific features by interpolation from the 4D feature space. It can be easily implemented by \textit{grid\_sample} method in PyTorch~\cite{pytorch}. In the following sections, the sampled spatio-temporal feature for an action proposal is denoted by $F \in \mathbb{R}^{T_{in}\times P_{in}\times D}$. 

\subsection{Spatio-temporal Decoupled Feature Mixing} \label{para:adamix}

After feature sampling, we perform spatio-temporal decoupled feature mixing under the guide of spatial queries and temporal queries to enhance the instance-specific representation. The prior information contained in the queries is utilized to guide the weights attached to different feature channels and points. We first describe the feature mixing process in detail and then present the mixing strategy.

\noindent \bf Channel mixing. \rm Different from MLP-Mixer~\cite{mlpmixer}, our mixing parameters are generated adaptively. Given a query $q \in \mathbb{R}^{D}$ and its corresponding features $f \in \mathbb{R}^{P\times D}$ to be mixed, we first use a linear layer as parameter generator to generate query-specific channel-mixing weights $M_c \in \mathbb{R}^{D\times D}$, and then apply plain matrix multiplication on $f$ and $M_c$ to perform channel-mixing, given by:
\begin{equation}
    M_c = \text{Linear}(q) \in \mathbb{R}^{D\times D},
\end{equation}
\begin{equation}
    \text{CM}(f) = \text{ReLU}(\text{LayerNorm}(f \times M_c)).
\end{equation}
where LayerNorm stands for layer normalization~\cite{layernorm}. We use CM$(f)\in \mathbb{R}^{P \times D}$ to denote the channel-wise mixed output features.

\noindent \bf Point mixing. \rm After channel-mixing, we perform point-wise mixing in a similar way. Suppose $P^{\prime}$ is the number of point-wise mixing out patterns, we use PCM$(f)\in \mathbb{R}^{D\times P^{\prime}}$ to denote the point-wise mixing output features, given by:
\begin{equation}
    M_p = \text{Linear}(q) \in \mathbb{R}^{P\times P^{\prime}},
\end{equation}
\begin{equation}
    \text{PCM}(f) = \text{ReLU}(\text{LayerNorm}({\text{CM}(f)}^T \times M_p)).
\end{equation}

\noindent \bf Query update. \rm The final output PCM$(f)$ is flattened, linearly transformed to $D$ dimension, and added to the query $q$ to update it, given by:
\begin{equation}
    q^{\prime} = q + \text{Linear}(\text{Flatten}(\text{PCM}(f))).
\end{equation}

\noindent \bf Decoupled mixing strategy. \rm We abbreviate the whole process of adaptive mixing as:
\begin{equation}
    q^{\prime} = \text{AM}(q, f, P, P^{\prime}).
\end{equation}

For \textit{keyframe action detection}, we factorize the sampled features into spatial features and temporal features by pooling and perform completely symmetrical spatial and temporal mixing with spatial queries and temporal queries respectively, given by:

\begin{equation}
    \begin{cases}q^{\prime}_{s} = \text{AM}(q_s, \text{TP}(F), P_{in}, P_{out}),\\ 
    q^{\prime}_{t} = \text{AM}(q_t, \text{SP}(F), T_{in}, T_{out}),\end{cases}  
\end{equation}
where TP and SP denote temporal pooling and spatial pooling respectively and $P_{out}$ and $T_{out}$ denote spatial and temporal mixing out patterns respectively.

For \textit{action tubelet detection}, we perform parallel and independent spatial mixing in each frame with the corresponding spatial queries. Supposing $q_{s}$ is a spatial query of frame $t$, we denote its corresponding sampled feature as $F[t] \in \mathbb{R}^{P_{in}\times D}$, then:
\begin{equation}
    \begin{cases}q^{\prime}_{s} = \text{AM}(q_s, F[t], P_{in}, P_{out}),\\ 
    q^{\prime}_{t} = \text{AM}(q_t, \text{SP}(F), T_{in}, T_{out}).\end{cases}  
\end{equation}

\subsection{STMixer Action Detector}
STMixer adopts a unified action decoder for both actor localization and action classification. The decoder comprises $M$ stacked ASAM modules followed by a feed-forward network (FFN) for human scores prediction and a short-term or long-term classifier for action score prediction. In this section, we represent the structure of ASAM module and specify the outputs of the prediction heads for both keyframe action detector (STMixer-K) and action tubelet detector (STMixer-T).

\noindent \bf ASAM module. \rm \label{para:ASAM} We first perform self-attention~\cite{attention} on spatial queries $Q_s$ and temporal queries $Q_t$ to capture the relation information between different instances. For \textit{action tubelet detection}, we adopt tubelet self-attention~\cite{TubeR} for spatial queries, that is, perform self-attention on queries of each frame and on queries of each tubelet sequentially. Then we perform adaptive sampling from the 4D feature space and decoupled adaptive mixing on the sampled feature as described before. The spatial queries $Q_s$ and temporal queries $Q_t$ are updated with mixed features. An FFN is applied on updated spatial queries $Q_s^{\prime}$ to update the positional queries $Q_p$. The updated queries $Q_p^{\prime}$, $Q_s^{\prime}$ and $Q_t^{\prime}$ are used as inputs for the next ASAM module.

\noindent \bf Prediction heads. \rm The output human boxes $\hat{Y_{b}}$ are decoded from final positional queries $Q_p^{\prime}$ directly, $\hat{Y_{b}} \in \mathbb{R}^{N\times 4}$ for \textit{keyframe action detection} and $\hat{Y_{b}} \in \mathbb{R}^{N\times T\times 4}$ for \textit{tubelet detection}. For \textit{keyframe action detection}, we apply an FFN on $Q_s^{\prime}$ to predict human scores $\hat{Y_{h}} \in \mathbb{R}^{N\times 2}$ which indicates the confidence values that each box belongs to the human category and background. Based on the concatenation of spatial queries $Q_s^{\prime}$ and temporal queries $Q_t^{\prime}$, we use a short-term or long-term classifier to predict action scores $\hat{Y_{a}} \in \mathbb{R}^{N\times C}$, where $C$ is the number of action classes. For \textit{action tubelet detection}, we perform temporal average pooling on spatial queries $Q_s^{\prime}$ to aggregate the appearance information of different frames, then we concatenate them with their corresponding temporal queries $Q_t^{\prime}$. We use an FFN to predict classification scores $\hat{Y_{c}} \in \mathbb{R}^{N\times (C+1)}$, where the last dimension represents the probability of background. 

\noindent \bf Long-term classifier. \rm \label{para:longterm} For \textit{keyframe action detection} on the AVA dataset, beyond the short-term FFN classifier which predicts action scores based on short-term information of current queries, we also design a long-term classifier which refers to a query bank for long-term information. Our design of the long-term classifier is adapted from LFB~\cite{lfb}. We train an STMixer model without long-term information first. Then, for a video of $\mathcal{T}$ clips, we use the trained STMixer to perform inference on each clip of it. We store the concatenation of the spatial and temporal queries from the last ASAM module corresponding to the $k$ highest human scores in the query bank for each clip. We denote the stored queries for clip of time-step $t$ as $L_t\in \mathbb{R}^{k\times d}$ where $d=2D$, and the long-term query bank of the video as $L=[L_0,L_1,...,L_{\mathcal{T}-1}]$. Given the long-term query bank of all videos, we train an STMixer model with a long-term classifier from scratch. For video clip $t$, we first sample a window of length $w$ from the long-term query bank centered at it and stack this window into $\tilde{L_{t}}\in \mathbb{R}^{K\times d}$:
\begin{equation}
    \tilde{L_{t}} = \text{stack}([L_{t-w/2},..., L_{t+w/2-1} ]).
\end{equation}
We then infer current queries to $\tilde{L_{t}}$ for long-term information by cross-attention~\cite{attention}:
\begin{equation}
    S_t^\prime = \text{cross-attention}(S_t,\tilde{L_{t}}),
\end{equation}
where $S_t\in \mathbb{R}^{N\times d}$ is the concatenation of the output spatial queries and temporal queries of the last ASAM module. The output $S_t^\prime$ is then channel-wise concatenated with $S_t$ for action scores prediction.

\noindent \bf Differences between STMixer-K and STMixer-T. \rm
We outline the main differences between keyframe action detector STMixer-K and action tubelet detector STMixer-T here: 1) The input video clips are sampled in different ways. STMixer-K samples video clips centered at each keyframe with a stride to cover a longer temporal range while STMixer-T samples consecutive frames for dense box prediction. 2) STMixer-K uses temporal downsampling in the backbone for temporal information aggregation while STMixer-T removes temporal downsampling as frame-specific information is required for box prediction in each frame. 3) The spatial queries $Q_s$ and positional queries $Q_p$ in STMixer-T have a temporal dimension as we need to predict a sequence of temporally continuous boxes for each tubelet. 4) STMixer-K calculates sampling points in the keyframe and then propagates them along the temporal axis while STMixer-T calculates sampling points in each frame based on the corresponding spatial query and positional query. 5) The label assignment and loss calculation in training are based on keyframe detection results and tubelets for STMixer-K and STMixer-T respectively. In summary, STMixer-T and STMixer-K are tailored for keyframe action detection and action tubelet detection respectively. Despite sharing the core designs of 4D feature space and adaptive feature sampling and mixing mechanism, the specific implementations of STMixer-K and STMixer-T are very different because of the two different forms of action detection task.

\subsection{Training}

For query-based detection methods that formulate the detection task as a set prediction problem, we need to assign ground-truth labels to predictions for loss calculating and model training. Following the practice of former query-based detectors~\cite{DETR,deformableDETR}, we first use the Hungarian algorithm to find an optimal bipartite matching $\sigma_*(\cdot)$ between the predictions and ground truths based on a matching loss $\mathcal{L}_{m}$. Then we calculate training loss $\mathcal{L}$ based on the optimal matching $\sigma_*(\cdot)$.

For \textit{keyframe action detection}, we calculate $\mathcal{L}_{m}$ and $\mathcal{L}$ as follows:
\begin{equation}
\begin{split}
    \mathcal{L}_{m}(\sigma)=\sum_{i=1}^{N}\lambda_{1}\mathcal{L}_{cls}(\hat{Y^{i}_{h}},Y^{\sigma(i)}_{h})+\lambda_{2}\mathds{1}_{\sigma(i)\ne \emptyset }\mathcal{L}_{L_1}(\hat{Y^{i}_{b}},Y^{\sigma(i)}_{b}) \\  +\lambda_{3}\mathds{1}_{\sigma(i)\ne \emptyset }\mathcal{L}_{giou}(\hat{Y^{i}_{b}},Y^{\sigma(i)}_{b}), \IEEEeqnarraynumspace \IEEEeqnarraynumspace 
\end{split}
\end{equation}
\begin{equation}
\begin{split}
    \mathcal{L}= \mathcal{L}_{m}(\sigma_*) +\sum_{i=1}^{N}\lambda_{4}\mathds{1}_{\sigma_*(i)\ne \emptyset }\mathcal{L}_{act}(\hat{Y^{i}_{a}},Y^{\sigma_*(i)}_{a}),
\end{split}
\end{equation}
where $\hat{Y}$ and $Y$ denote the predictions and the ground truths respectively, and $\sigma(\cdot)$ is a permutation of the prediction to match with the ground truths. $\mathcal{L}_{cls}$ denotes the cross-entropy loss over two classes (human and background). $\mathcal{L}_{L_1}$ and $ \mathcal{L}_{giou}$ are box loss inherited from \cite{DETR,deformableDETR,sparsercnn}. $\mathcal{L}_{act}$ is binary cross entropy loss for action classification. The final training loss $\mathcal{L}$ is calculated based on the optimal bipartite matching $\sigma_*(\cdot)$ which minimizes $\mathcal{L}_{m}$. $\lambda_{1}$, $\lambda_{2}$, $\lambda_{3}$, and $\lambda_{4}$ are corresponding weights of each term.

For \textit{action tubelet detection}, $\mathcal{L}_{m}$ and $\mathcal{L}$ are calculated as follows:
\begin{equation}
\begin{split}
    \mathcal{L}_{m}(\sigma)=\sum_{i=1}^{N}\lambda_{1}\mathcal{L}_{cls}(\hat{Y^{i}_{c}},Y^{\sigma(i)}_{c})+\lambda_{2}\mathds{1}_{\sigma(i)\ne \emptyset }\mathcal{L}_{L_1}(\hat{Y^{i}_{b}},Y^{\sigma(i)}_{b}) \\  +\lambda_{3}\mathds{1}_{\sigma(i)\ne \emptyset }\mathcal{L}_{giou}(\hat{Y^{i}_{b}},Y^{\sigma(i)}_{b}), \IEEEeqnarraynumspace \IEEEeqnarraynumspace 
\end{split}
\end{equation}
\begin{equation}
\begin{split}
    \mathcal{L}= \mathcal{L}_{m}(\sigma_*),
\end{split}
\end{equation}
where $\mathcal{L}_{cls}$ is focal loss for action classification over $c+1$ classes.

\section{Experiments}

\subsection{Datasets and Evaluation Metrics}

We evaluate our STMixer framework on five action detection benchmarks, including AVA and AVA-Kenitics with sparse annotations on keyframes and UCF101-24, JHMDB51-21 and MultiSports with dense action tube annotations. We first give a brief introduction to these datasets and then specify the evaluation metrics we use.

\noindent \bf Datasets for keyframe action detection. \rm 

AVA~\cite{ava} contains 211k keyframes annotated with frame-level action instances for training and 57k for validation. The keyframes are selected at 1FPS from 430 15-minute movie videos. There are 80 action classes in AVA including human poses, human-object interactions, and human-human interactions. Each person is annotated with one or more action classes, thus action detection on AVA is a multi-label detection task. Following the standard evaluation protocol~\cite{ava}, we report our results on 60 action classes that have at least 25 validation examples.

AVA-Kinetics~\cite{avakinetics} is a superset of AVA. It annotates frame-level action instances on a part of Kinetics-700 video clips following AVA's annotation protocol. Only one keyframe is annotated for each selected video clip. The introduction of Kinetics-700 videos makes the dataset not limited to movie videos, and the scenes contained become more diverse. It also alleviates the problem of too few instances in some classes. In total, AVA-Kinetics extends AVA by about 140K keyframes in the training set and 32K keyframes in the validation set. 

\noindent \bf Datasets for action tube detection. \rm 

UCF101-24~\cite{ucf101} is a subset of UCF101. It contains 3,207 untrimmed videos annotated with video-level action tubes of 24 classes. As the common setting, we report the performance on the first split. 

JHMDB51-21~\cite{jhmdb} consists of 928 temporally trimmed videos from 21 action classes. Following the common setting, results averaged over three splits are reported.  

MultiSports~\cite{multisports} is a large-scale benchmark for video-level action tube detection. It contains 1,574 video clips for training and 555 for validation. There are 66 action classes from sports of football, basketball, volleyball, and aerobic gymnastics. In total, there are 18,422 action tube instances in the training set and 6,577 in the validation set. Following the official evaluation guide, we evaluate our method on 60 classes that have at least 25 instances in the validation set.

\noindent \bf Evaluation metrics. \rm For keyframe action detection on AVA and AVA-Kinetics, we take Frame mean Average Precision at an IoU threshold of 0.5 (Frame mAP@0.5) as the evaluation metric. For UCF101-24, JHMDB51-21 and MultiSports, both Frame mAP and Video mAP are reported. Video mAP is calculated under different 3D IoU thresholds. 3D IoU is defined as the temporal domain IoU of two tubes, multiplied by the average of the spatial IoU between the overlapped frames. Video mAP is a more comprehensive metric that can better evaluate the accuracy of both spatial and temporal localization.

\subsection{Implementation Details}
\noindent \bf Network configurations. \rm We configure the dimension $D$ of both spatial and temporal queries to 256 and set the number of both queries $N$ equaling 100. The number of sampling points $P_{in}$ in each temporal frame is set to 32. The spatial and temporal mixing out patterns $P_{out}$ and $T_{out}$ are set to 4 times the number of sampling points $P_{in}$ and temporal length of input feature maps $T_{in}$ respectively. Following multi-head attention~\cite{attention} and group convolution~\cite{groupconv}, we split the channel $D$ into 4 groups and perform group-wise sampling and mixing. We stack 6 ASAM modules for keyframe action detection on AVA and AVA-Kinetics and 3 for action tubelet detection on UCF101-24, JHMDB51-21 and MultiSports. For the long-term classifier for AVA, we set the number of stored queries of each clip $k$ as 5 and window length $w$ as 60. The number of cross-attention layers in the long-term classifier is set to 3.

\noindent \bf Losses and optimizers. \rm We set the loss weights for keyframe action detection as $\lambda_{1} = 2.0$, $\lambda_{2} = 2.0$, $\lambda_{3} = 2.0$ and $\lambda_{4} = 24.0$ and for action tubelet detection as $\lambda_{1} = 2.0$, $\lambda_{2} = 2.0$ and $\lambda_{3} = 2.0$. We use AdamW~\cite{adamw} optimizer with a weight decay rate of $1 \times 10^{-4}$. Following \cite{DETR,woo}, intermediate supervision is applied after each ASAM module. 

\TableAbFeatureSpace

\TableAbFeatureSampling

\noindent \bf Training and inference recipes. \rm For AVA and AVA-Kinetics, we train STMixer detectors for 10 epochs with an initial learning rate of $2.0 \times 10^{-5}$ and batchsize of 16. The learning rate and batchsize can be tuned according to the linear scaling rule~\cite{linearscale}. We randomly scale the short size of the training video clips to 256 or 320 pixels. Color jittering and random horizontal flipping are also adopted for data augmentation.

For UCF101-24 and MultiSports, the same initial learning rate and batchsize as AVA are used. The model is trained for 30K and 120K iterations for UCF101-24 and MultiSports respectively. For UCF101-24, we scale the short side of all training videos to 256 while for MultiSports, we randomly scale the short size of the training video clips to 256 or 288 pixels. Other data augmentations remain the same with AVA. For JHMDB, we train the model for 20k iterations with an initial learning rate of $5.0 \times 10^{-6}$.

For inference, we scale the short size of input frames to 256 as the common setting unless otherwise specified. For each prediction, if the predicted probability that it belongs to the background category is lower than the preset threshold, we take it as a detection result. We set the threshold to 0.3 for AVA, AVA-kinetics, and JHMDB51-21, and 0.7 for UCF101-24 and MultiSports. For UCF101-24, JHMDB51-21 and MultiSports, we need to link the action tubelets into video-level action tubes. The same linking strategy as the former methods~\cite{act,moc,TubeR} is adopted, which is proposed by ACT~\cite{act}.

\TableAbFeatureMixing

\TableAbOutPatterns

\begin{figure}[!t]
    \centering
    \includegraphics[width=7.5cm]{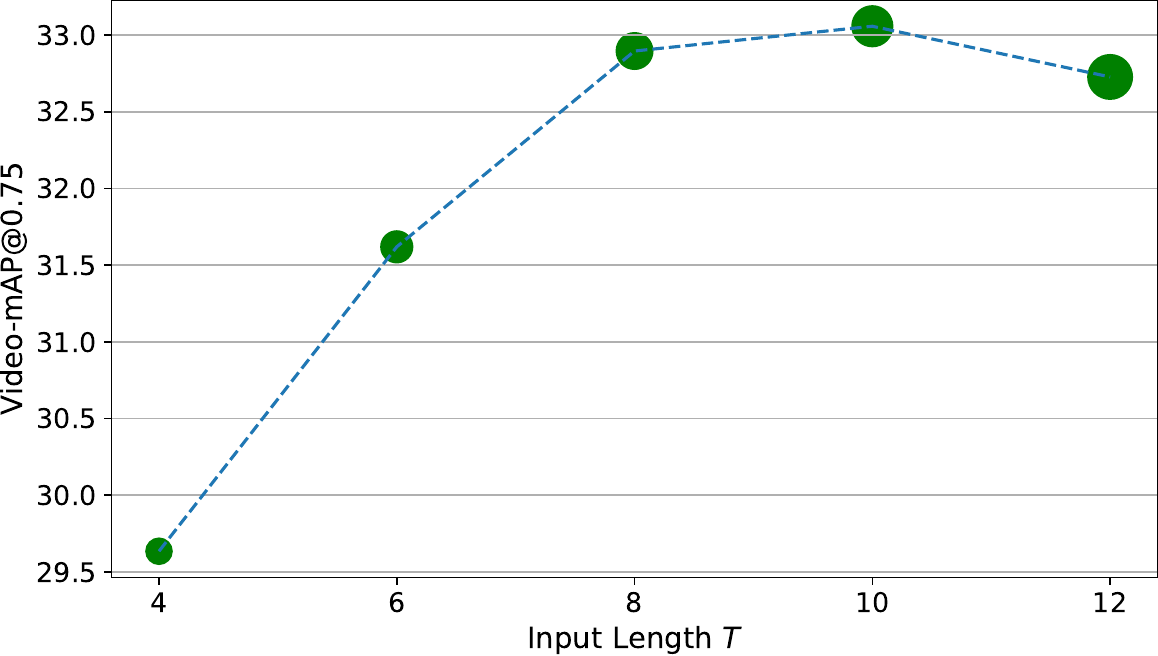}
    \caption{Ablations on input length $T$ for tubelet detection on UCF101-24. The size of the green circle represents the number of GFLOPs. Setting $T$ around 8 to 10 works well.}
    \label{fig:length}
\end{figure}

\subsection{Ablation Study} 
\label{subsection:ablation}
We conduct ablation experiments on AVA v2.2 and UCF101-24 to investigate the influence of core components in STMixer-K and STMixer-T. STMixer-K is evaluated on AVA v2.2 and STMixer-T is evaluated on UCF101-24. For AVA v2.2, the SlowOnly ResNet-50 backbone~\cite{fan2020pyslowfast} is used for video feature extraction and each input video clip has 4 frames with a sampling rate of 16. For UCF101-24, the CSN-152 backbone is used and each input clip consists of 8 consecutive frames unless otherwise specified. We report both detection mAP and GFLOPs for effectiveness and efficiency comparison. The default choices are colored in \colorbox[HTML]{EAEAEA}{gray}.

\noindent \bf Ablations on 4D feature space. \rm We first investigate two different ways for the 4D feature space construction. As shown in Table~\ref{tab:abfeaturespace}, constructing 4D feature space by simple lateral $1\times 1\times 1$ convolution achieves comparable detection accuracy while being more efficient than using full FPN~\cite{fpn} with a top-down pathway. Then we test using feature maps of different numbers of scales. Using feature maps of all 4 different scales achieves the best results on both AVA v2.2 and UCF101-24. Compared with only using feature maps of the last stage, Frame mAP@0.5 on AVA v2.2 is improved by 1.0 points and Video mAP@0.75 on UCF101-24 is improved by 1.6 points. These results demonstrate the importance of multi-scale information in action detection and the effectiveness of our design of the 4D feature space.

\noindent \bf Ablations on feature sampling. \rm In Tabel~\ref{tab:abfeaturesampling}, we investigate the influence of the feature sampling strategy and the number of sampling points. We compare 2 different feature sampling strategies. For fixed grid feature sampling, we test sampling $6\times6$ or $7\times7$ feature points inside each actor box by interpolation, which is actually the RoIAlign~\cite{maskrcnn} operation adopted by many former methods~\cite{woo,vat}. Compared with fixed grid sampling, our query-based adaptive sampling achieves significantly better detection performance. Adaptive sampling 32 points per group outperforms fixed grid sampling 49 points by 1.2 Frame mAP@0.5 on AVA v2.2 and 1.2 Video mAP@0.75 points on UCF101-24. The results show that the fixed RoIAlign operation for feature sampling fails to capture useful context information outside the box while our adaptive sampling strategy enables the detector to mine discriminative features from the whole 4D feature space. The performance gap between fixed grid sampling and our adaptive sampling is larger on AVA v2.2 than UCF101-24. This is because AVA pays more attention to the interaction between the actor and context and the importance of the ability of modeling context is more prominent on AVA. In Table~\ref{tab:abfeaturesampling}, we further investigate the influence of the number of sampling points per group $P_{in}$ in each temporal frame. Setting $P_{in}=32$ achieves the best detection performance on both datasets.

\TableAVA

\TableAK

\noindent \bf Ablations on feature mixing. \rm In Table~\ref{tab:abfeaturemixing}, we compare different feature mixing strategies. We first demonstrate the benefit of query-guided adaptive mixing. Frame mAP@0.5 on AVA v2.2 drops by 0.6 and Video mAP@0.75 on UCF101-24 drops by 1.9 when using fixed parameter mixing. For adaptive mixing, the mixing parameters are dynamically generated based on a specific query, thus more powerful to enhance the presentation of each specific action instance proposal. We further compare different adaptive feature mixing strategies. From the results in Table~\ref{tab:abfeaturemixing}, it is demonstrated that both spatial appearance and temporal motion information are important for action detection. However, coupled spatio-temporal feature mixing using queries of a single type has a high computational complexity. Decoupling features along spatial and temporal dimensions saves computing costs. Our decoupled parallel spatio-temporal feature mixing outperforms sequential spatio-temporal feature mixing by 0.5 Frame mAP@0.5 on AVA v2.2 and 1.6 Video mAP@0.75 on UCF101-24. This is because actor localization at keyframes only needs spatial appearance information and parallel mixing will reduce the effect of temporal information on localization. Also, by concatenating temporal queries to spatial queries, more temporal information is leveraged for action classification. In Table~\ref{tab:aboutpatterns}, we investigate different spatial and temporal mixing out patterns $P_{out}$ and $T_{out}$ 2 times to 6 times the number of sampling points $P_{in}$ and temporal length of input feature maps $T_{in}$. Setting $P_{out}$ and $T_{out}$ equaling 4 times $P_{in}$ and $P_{out}$ achieves the best performance. 

\noindent \bf Ablations on input length. \rm We investigate the influence of input length $T$ on action tube detention on UCF101-24 in Fig.~\ref{fig:length}. Benefitting from richer temporal context and motion information, the detection performance increases as we increase $T$ from 2 to 10. However, as we continue to increase $T$, the detection performance degrades. This is because action tubelet detectors assume that the length of all action tubes is greater than $T$ frames. When $T$ is set too large, this assumption will cause some action tubes shorter than $T$ frames to never be detected.

\TableTube

\TableMultiSports

\subsection{Comparison with the State-of-the-Art Methods}

\noindent \bf AVA and AVA-Kinetics. \rm We compare our proposed STMixer-K keyframe action detector with state-of-the-art methods on AVA v2.1 and v2.2 in Table~\ref{tab:ava}. We first compare STMixer-K to methods using an extra offline human detector. Our STMixer-K with SlowFast-R101 backbone achieves 30.6 and 30.9 mAP when not using long-term features. With long-term query support, our STMixer-K reaches 32.6 and 32.9 mAP on AVA v2.1 and v2.2 respectively. To demonstrate the generalization ability of our method, we conduct experiments with ViT~\cite{ViT} backbone. Compared with the two-stage counterparts, STMixer brings performance improvements while eliminating dependence on an extra detector. Although ViT is considered to have a global receptive field, our adaptive sampling and decoding mechanism could serve as a supplement to improve the flexibility of the model. Compared to previous end-to-end methods, our STMixer-K achieves the best results. Our STMixer-K outperforms WOO~\cite{woo} by 2.0 and 1.7 mAP even though WOO test models at 320 resolution. STMixer-K also consistently outperforms TubeR~\cite{TubeR} on AVA v2.1 or v2.2, using or not using long-term features. 

\begin{figure*}[!t]
    \centering
    \includegraphics[width=\textwidth]{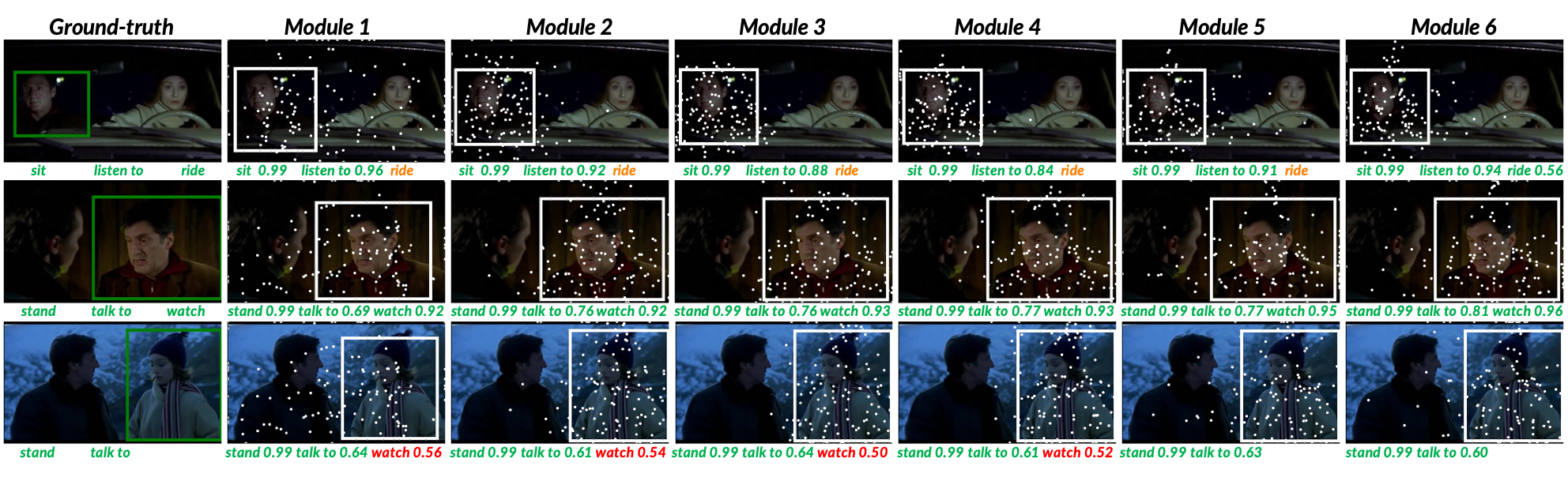}
    \vspace{-7mm}
    \caption{Visualization of sampling points and detection results (STMixer-K) on AVA. We display the actor bounding boxes and action classes of three ground-truth action instances in the first column. Each ASAM module's sampling points, predicted actor bounding boxes, and action scores are displayed in the following columns. The correctly predicted action classes are displayed in \textcolor[HTML]{23B051}{green}, missing in \textcolor{orange}{orange}, and wrongly predicted in \textcolor{red}{red}. Intuitively, STMixer mines discriminative context features outside the bounding box for better action detection.}
    \label{fig:vis}
\end{figure*}

\begin{figure*}[!t]
    \centering
    \includegraphics[width=\textwidth]{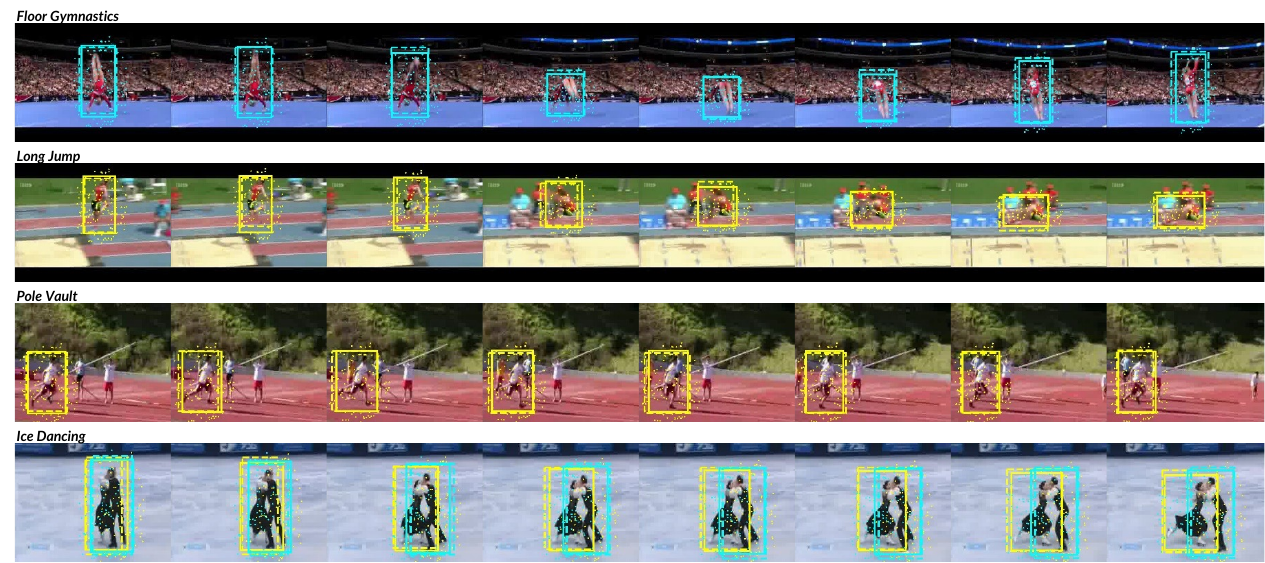}
    \caption{Visualization of sampling points and tubelet detection results (STMixer-T) on UCF101-24. Each row shows a video clip and the detected tubelets on it. The frames of each video clip are arranged in order from left to right. The bounding boxes and sampling points of the same actor are displayed in the same color. The dotted boxes are the ground-truth boxes, and the solid boxes are our predictions. In the first two rows, the actor's pose changes drastically in different frames. In the last two rows, the actor's tracking faces interference from the background or other actors. Our STMixer localizes and tracks the actor accurately in these difficult cases. }
    \label{fig:visucf}
\end{figure*}

We compare mAP versus GFLOPs on AVA v2.2 in Fig.~\ref{fig:flops} to show the efficiency of our STMixer-K. AIA~\cite{aia} and ACARN~\cite{acarn} do not report their GFLOPs. As they are built on SlowFast~\cite{slowfast} framework and also use an offline human detector but introduce extra modules to model interaction, SlowFast can serve as a lower bound of complexity for them. For a fair comparison, we report results for no long-term feature version of each method. As shown in Fig.~\ref{fig:flops}, due to an extra human detector being needed, SlowFast~\cite{slowfast}, CSN~\cite{csn}, and VideoMAE~\cite{vmae} have much higher GFLOPs than end-to-end methods with same backbone. Among end-to-end methods, STMixer-K achieves the best effectiveness and efficiency balance. STMixer-K outperforms WOO~\cite{woo} by 1.8 mAP while having much lower GFLOPs (135 versus 252). With a slight GFLOPs increase (126 versus 120), STMixer-K outperforms TubeR~\cite{TubeR} by 1.7 mAP.

We further evaluate our STMixer-K on AVA-Kinetics. Without bells and whistles, our STMixer-K achieves 37.5 mAP, surpassing ACARN~\cite{acarn} with test-time augmentation by 1.1 points and Co-finetuning~\cite{coft} using extra training data by 1.3 points.

\noindent \bf UCF101-24 and JHMDB51-21. \rm We compare our STMixer-T with state-of-the-art methods on UCF101-24 and JHMDB51-21 in Table~\ref{tab:tube}. Following TubeR, we adopt CSN-152 as the backbone for feature extraction, but we take video clips composed of consecutive frames as inputs for dense prediction on each frame following most former tubelet detectors. On UCF101-24, our STMixer-T achieves new state-of-the-art results, surpassing the previous methods by a larger margin. The videos in UCF101-24 are untrimmed. For accurate localization of the temporal boundaries of action tubes, it is required to utilize the temporal information more effectively to determine whether an action is occurring in the current clip. Former methods rely on optical flow for temporal modeling, however, the extraction of optical flow is time-consuming. Our STMixer-T exhibits excellent temporal modeling capabilities, surpassing these two-stream methods significantly with only RGB frames as input. Compared with RGB-only models, the performance gap is even larger. Our STMixer-T outperforms MOC by 12.4 points on Frame mAP@0.5 and 14.7 points on Video mAP@0.5. Compared with TubeR, using the same CSN-152 backbone, STMixer-T outperforms it by an evident margin on all metrics though it uses a longer temporal context. JHMDB51-21 is a small dataset with trimmed videos. Our STMixer-T achieves 87.0 on Video mAP@0.5, outperforming TubeR by 4.7 points and CFAD with optical flow by 1.7 points. This result demonstrates that STMixer-T can also learn effective feature sampling and decoding on a small dataset, and because of its simplicity, it can also alleviate overfitting.

\begin{figure*}[!t]
\centering
\subfloat[Visualizations of action tubes on UCF101-24. We sample 10 frames from each video at equal intervals for visualization.]{\includegraphics[width=\textwidth]{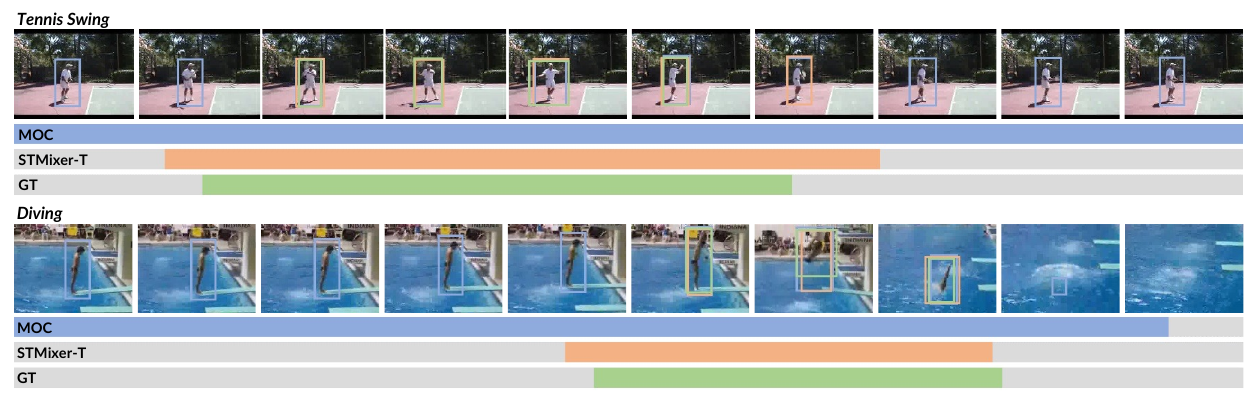}
\label{fig:tubeucf}}
\hspace{0.5em}
\subfloat[Visualizations of action tubes on MultiSports. We sample 10 frames at equal intervals from the union of the temporal ranges of the three action tubes for visualization, which is marked by a red box. ]{\includegraphics[width=\textwidth]{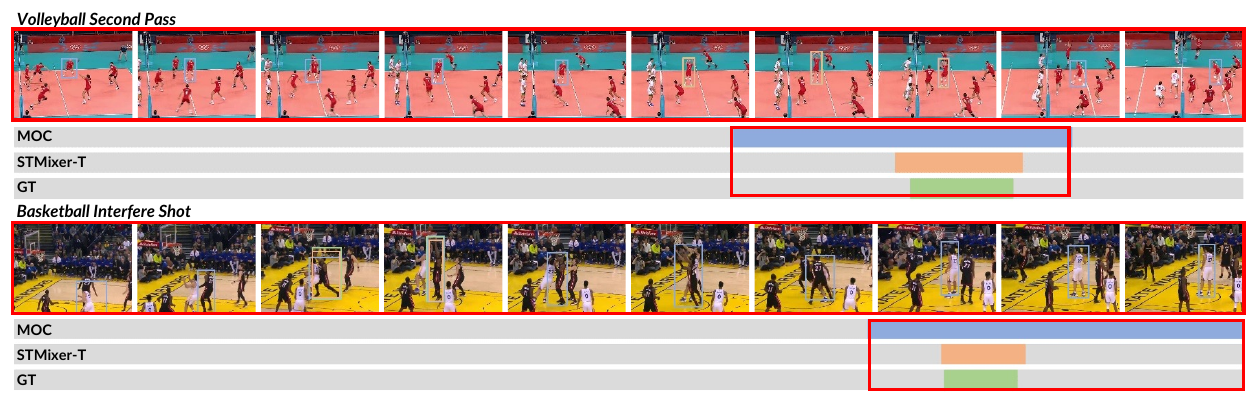}\label{fig:tubemul}}
\caption{Visualizations of action tubes detected by STMixer-T on (a) UCF101-24; (b) MultiSports. We take MOC~\cite{moc} for an comparison. MOC tends to produce much longer action tubes when the appearances of different frames are similar. Our STMixer-T can give more accurate temporal boundaries because it has a stronger temporal modeling capability to better judge the start and end of actions. }
\label{fig:tube}
\end{figure*}

In Table~\ref{tab:tube}, the results of STMixer-K are also provided. STMixer-K achieves a very competitive Frame-mAP a z it is specifically designed for keyframe action detection. However, on the metric of Video-mAP that better reflects the performance of the video-level action tube detection, STMixer-K lags behind STMixer-T, and the gap is even larger when the IoU threshold is higher.

\noindent \bf MultiSports. \rm The results on the MultiSports dataset are reported in Table~\ref{tab:multisports}. TAAD~\cite{taad} uses extra tracking results generated by an off-the-shelf tracking model for feature aggregation and tube linking. Compared with TAAD, the performance of our STMixer-T is still competitive, with the average Video mAP over IoU thresholds from 0.5 to 0.95 only 1.7 points lower. When slightly improving the resolution to $288\times512$ for inference, our STMixer-T achieves an average Video mAP of 33.7, which is comparable to TAAD using an extra tracker. As for fair comparison, our STMixer-T outperforms TAAD without an extra tracker by 3.9 points on Video mAP@0.5 and 4.8 points on average Video mAP. Compared with PCCA~\cite{pcca}, the performance gap is even larger. Our STMixer-T surpasses PCCA by 18.3 points on Video mAP@0.5 and 12.9 points on average Video mAP with input clips of shorter length and the same resolution.

\section{Model Visualization}

\subsection{Visualization of Keyframe Action Detection}
For keyframe action detection on AVA v2.2, we provide the visualization of sampling points and detection results of each ASAM module in STMixer-K in order in Fig.~\ref{fig:vis}. In the first few ASAM modules, the sampling points are quickly concentrated on the action performer, and the actor localization accuracy gets improved rapidly. In the following ASAM modules, some of the sampling points get out of the human box and spread to the context. Benefiting from the gathered context and interaction information, the action recognition accuracy gets improved while localization accuracy is not compromised. In Fig.~\ref{fig:vis}, we show clear improvements in three aspects: predicting missing classes (\textcolor{orange}{ride} in row 1),  improving the confidence score of correctly predicted classes (\textcolor[HTML]{23B051}{talk to} in row 2), and removing wrongly predicted classes (\textcolor{red}{watch} in row 3). To recognize an action ``ride'', we need to observe the person is in a car and someone is driving. For recognition of ``talk to", we need to know if a person is listening in the context, and for ``watch", we need to know if a person is in the actor's sight. The improvements are mostly related to these classes of interaction, which indicates our STMixer is capable of mining discriminative interaction information from the context.

\subsection{Visualization of Action Tube Detection}

We first demonstrate the accuracy of spatial localization and tracking in each tubelet through the visualization of sampling points and bounding boxes of several detected tubelets in Fig.~\ref{fig:visucf}. In the first and second rows, the poses of the actors in different frames vary greatly. Our STMixer-T gives accurate actor bounding boxes in each frame, which shows the rationality and effectiveness of our independent feature sampling and spatial mixing in the feature space of each frame. In the third row, there are several people not related to the action but very close to the action performer. Our STMixer-T accurately distinguishes and tracks the real action performer. In the last row, the two action performers are very close and are exchanging positions quickly. Our STMixer-T still accurately tracks each of them. 

We further provide the temporal range of the final linked action tubes in Fig.~\ref{fig:tube} to demonstrate the accuracy of temporal boundary judgment. We take a strong tubelet detector MOC for comparison. MOC relies more on appearance information and lacks temporal modeling. When the appearance of different video frames is similar, it is difficult for MOC to distinguish the background segments, so it tends to predict over-long action tubes. Benefiting from the stronger temporal modeling ability brought by our proposed temporal mixing, our STMixer-T can accurately judge whether an action occurs in each video clip so that action tubes with accurate temporal ranges can be obtained through the linking algorithm.

\section{Conclusion and Future Work}
In this paper, we have presented a new one-stage sparse action detector in videos, termed STMixer. Our STMixer yields a simple, neat, and efficient end-to-end action detection framework that can be implemented for both keyframe action detection and action tubelet detection. The core design of our STMixer is a set of learnable queries to decode all action instances in parallel. The decoding process is composed of an adaptive feature sampling module to identify important features from the entire spatio-temporal domain of video, and an adaptive feature mixing module to dynamically extract discriminative representations for action detection. Our STMixer achieves a new state-of-the-art performance on five challenging benchmarks of AVA, AVA-Kinetics, UCF101-24, JHMDB51-21 and MultiSports with less computational cost than previous end-to-end methods. We hope STMixer can serve as a strong baseline for future research on both keyframe action detection and action tube detection.

In the future, we would like to consider better designs in our dynamic sampling and mixing mechanism to further improve its efficiency in both parameters and computation.

\section*{\bf Acknowledgement} 
This work is supported by the National Key R$\&$D Program of China (No. 2022ZD0160900), the National Natural Science Foundation of China (No. 62076119, No. 61921006), the Fundamental Research Funds for the Central Universities (No. 020214380099), and the Collaborative Innovation Center of Novel Software Technology and Industrialization.

\ifCLASSOPTIONcaptionsoff
  \newpage
\fi

{
\bibliographystyle{IEEEtran}
\bibliography{egbib}
}

\end{document}

%% file: mytables.tex
\usepackage{multirow, overpic, textpos, array}
\usepackage{makecell}
\usepackage{footnote}
\usepackage[table]{xcolor}
\usepackage{colortbl}

\usepackage{wrapfig,lipsum,booktabs}
\newlength\savewidth

\newcolumntype{x}[1]{>{\centering\arraybackslash}p{#1pt}}
\newcolumntype{y}[1]{>{\raggedright\arraybackslash}p{#1pt}}
\newcolumntype{z}[1]{>{\raggedleft\arraybackslash}p{#1pt}}

\usepackage{xcolor, colortbl}
\definecolor{Gray}{gray}{0.9}
\definecolor{baselinecolor}{gray}{0.9}

\newcolumntype{a}{>{\columncolor{Gray}}c}

\newcommand{\TableAbFeatureSpace}{
\begin{table}[!t]
    \centering
    \setlength{\tabcolsep}{3.0pt}
    \renewcommand{\arraystretch}{1.1}
    \caption{Ablations on 4D feature space. Simple lateral convolution works well and using feature maps of all 4 scales achieves the best performance. }
    {
   \begin{tabular}{cccccc}
\toprule
\multirow{2}{*}{Construction} & \multirow{2}{*}{Scales} & \multicolumn{2}{c}{AVA v2.2 (STMixer-K)} & \multicolumn{2}{c}{UCF101-24 (STMixer-T)} \\ \cmidrule(lr{0.5em}){3-4}\cmidrule(l{0.5em}r{0.5em}){5-6}
 &  & F-mAP@0.5 & GFLOPs & V-mAP@0.75 & GFLOPs \\ \midrule
Full FPN & 4 & 22.9 & 45.8 & 32.4 & 149.8 \\ 
\hline
\multirow{4}{*}{Lateral Conv.} & 4 & \cellcolor[HTML]{EAEAEA}23.1 & \cellcolor[HTML]{EAEAEA}44.4 & \cellcolor[HTML]{EAEAEA}32.8 & \cellcolor[HTML]{EAEAEA}148.4 
\\
 & 3 & 23.0 & 43.3 & 32.8 & 145.5 \\
 & 2 & 22.7 & 42.7 & 32.6 & 144.1 \\
 & 1 & 22.1 & 42.5 & 31.2 & 143.4 \\ \bottomrule
\end{tabular}
   }
    \label{tab:abfeaturespace}
 \end{table}
}

\newcommand{\TableAbFeatureSampling}{
\begin{table}[!t]
    \centering
    \setlength{\tabcolsep}{4.5pt}
    \renewcommand{\arraystretch}{1.1}
    \caption{Ablations on feature sampling. Adaptive sampling performs better than fixed grid sampling even when sampling fewer points.}
    {
   \begin{tabular}{cccccc}
\toprule
\multirow{2}{*}{Sampling} & \multirow{2}{*}{$P_{in}$} & \multicolumn{2}{c}{AVA v2.2 (STMixer-K)} & \multicolumn{2}{c}{UCF101-24 (STMixer-T)} \\ \cmidrule(lr{0.5em}){3-4}\cmidrule(l{0.5em}r{0.5em}){5-6}
 &  & F-mAP@0.5 & GFLOPs & V-mAP@0.75 & GFLOPs \\ \midrule
\multirow{2}{*}{Fixed Grid} & 36 & 21.3 & 44.6 & 31.2 & 149.5 \\
 & 49 & 21.9 & 45.6 & 31.6 & 153.5 \\ \hline
\multirow{4}{*}{Adaptive} & 8 & 22.4 & 42.4 & 31.2 & 140.7 \\
 & 16 & 22.5 & 43.1 & 31.8 & 143.3 \\
 & 32 & \cellcolor[HTML]{EAEAEA}23.1 & \cellcolor[HTML]{EAEAEA}44.4 & \cellcolor[HTML]{EAEAEA}32.8 & \cellcolor[HTML]{EAEAEA}148.4 \\
 & 48 & 22.9 & 45.6 & 32.0 & 153.6 \\ \bottomrule
\end{tabular}
   }
    \label{tab:abfeaturesampling}
 \end{table}
}

\newcommand{\TableAbFeatureMixing}{
\begin{table}[!t]
    \centering
    \setlength{\tabcolsep}{2.6pt}
    \renewcommand{\arraystretch}{1.1}
    \caption{Ablations on feature mixing. Query-guided adaptive feature mixing outperforms fixed parameter mixing and our decoupled spatio-temporal feature mixing strategy works the best.}
    {
  \begin{tabular}{ccccccc}
\toprule
\multirow{2}{*}{Param.} & \multirow{2}{*}{Spa.} & \multirow{2}{*}{Tem.} & \multicolumn{2}{c}{AVA v2.2 (STMixer-K)} & \multicolumn{2}{c}{UCF101-24 (STMixer-T)} \\ \cmidrule(lr{0.5em}){4-5}\cmidrule(l{0.5em}r{0.5em}){6-7} 
 &  &  & F-mAP@0.5 & GFLOPs & V-mAP@0.75 & GFLOPs \\ \midrule
Fixed & \cmark & \cmark & 22.5 & 36.8 & 30.9 & 127.0 \\ \hline
\multirow{5}{*}{Generated} & \cmark & \cmark & \cellcolor[HTML]{EAEAEA}23.1 & \cellcolor[HTML]{EAEAEA}44.4 & \cellcolor[HTML]{EAEAEA}32.8 & \cellcolor[HTML]{EAEAEA}148.4 \\
 & \cmark & \xmark & 22.4 & 40.4 & 31.6 & 146.1 \\
 & \xmark & \cmark & 22.1 & 33.5 & 24.9 & 121.6 \\
 & \multicolumn{2}{c}{Sequential} & 22.6 & 43.6 & 31.2 & 161.8 \\
 & \multicolumn{2}{c}{Coupled} & 22.8 & 93.2 & 32.4 & 156.8 \\ \bottomrule
\end{tabular}
   }
    \label{tab:abfeaturemixing}
 \end{table}
}

\newcommand{\TableAbOutPatterns}{
\begin{table}[!t]
    \centering
    \setlength{\tabcolsep}{6pt}
    \renewcommand{\arraystretch}{1.1}
    \caption{Ablations on the number of mixing out patterns. Setting mixing out patterns to 4 times the input points works the best.}
    {
  \begin{tabular}{ccccc}
\toprule
\multirow{2}{*}{Out Patterns} & \multicolumn{2}{c}{AVA v2.2 (STMixer-K)} & \multicolumn{2}{c}{UCF101-24 (STMixer-T)} \\ \cmidrule(lr{0.5em}){2-3}\cmidrule(l{0.5em}r{0.5em}){4-5} 
 & F-mAP@0.5 & GFLOPs & V-mAP@0.75 & GFLOPs \\ \midrule
$\times 2$ & 22.5 & 40.2 & 30.4 & 133.0  \\
$\times 3$ & 22.8 & 42.3 & 31.9 & 140.7  \\
$\times 4$ & \cellcolor[HTML]{EAEAEA}23.1 & \cellcolor[HTML]{EAEAEA}44.4 & \cellcolor[HTML]{EAEAEA}32.8 & \cellcolor[HTML]{EAEAEA}148.4  \\
$\times 5$ & 22.9 & 46.4 & 32.5 & 156.2  \\
$\times 6$ & 22.6 & 48.4 & 31.4 & 163.9  \\ \bottomrule
\end{tabular}
   }
    \label{tab:aboutpatterns}
 \end{table}
}

\newcommand{\TableAVA}{
\begin{table*}[!t]
    \centering
    \setlength{\tabcolsep}{11.5pt}
    \renewcommand{\arraystretch}{1.1}
    \caption{Comparisons with state-of-the-arts on validation sets of AVA v2.1 and v2.2. \cmark of column ``Detector" denotes an extra human detector Faster RCNN-R101-FPN~\cite{fasterrcnn} is used. \cmark of column ``LF" denotes long-term features are used.}
    {
  \begin{tabular}{lcccccccc}
\toprule
\multicolumn{1}{c}{}  &         &             &              &            &        & & \multicolumn{2}{c}{\textbf{mAP}} \\ \cmidrule(l{0.5em}r{0.5em}){8-9} 
\multicolumn{1}{c}{\multirow{-2}{*}{\textbf{Method}}} &
  \multirow{-2}{*}{\textbf{Detector}} &
  \multirow{-2}{*}{\textbf{One-stage}} &
  \multirow{-2}{*}{\textbf{Input}} &
  \multirow{-2}{*}{\textbf{Backbone}} &
  \multirow{-2}{*}{\textbf{Pre-train}} &
  \multirow{-2}{*}{\textbf{LF}} &
  \textbf{v2.1} &
  \textbf{v2.2} \\ \midrule
\multicolumn{8}{l}{\textbf{Compare to methods with an extra human detector}} \\ \midrule
SlowFast~\cite{slowfast}              & \cmark & \xmark  & 32$\times$2  & SF-R101-NL   & K600       & \xmark & 28.2            & 29.0             \\
LFB~\cite{lfb}                   & \cmark & \xmark & 32$\times$2  & I3D-R101-NL  & K400       & \cmark & 27.7            & -              \\
CA-RCNN~\cite{carcnn}               & \cmark  & \xmark& 32$\times$2  & R50-NL       & K400       & \cmark & 28.0            & -              \\
AIA~\cite{aia}                   & \cmark  & \xmark& 32$\times$2  & SF-R101      & K700       & \cmark & 31.2            & 32.3           \\
ACARN~\cite{acarn}                 & \cmark & \xmark & 32$\times$2  & SF-R101      & K400       & \cmark & 30.0            & -              \\
ACARN~\cite{acarn}                 & \cmark & \xmark & 32$\times$2  & SF-R101-NL   & K600       & \cmark & -               & 31.4           \\
VideoMAE~\cite{vmae}              & \cmark  & \xmark& 16$\times$4  & ViT-B        & K400       & \xmark & -               & 31.8           \\
VideoMAE~\cite{vmae}              & \cmark  & \xmark& 16$\times$4  & ViT-L        & K700       & \xmark & -               & 39.3           \\ \hline
STMixer-K                  & \xmark  & \cmark& 32$\times$2  & SF-R101      & K700       & \xmark & 30.6            & 30.9           \\
STMixer-K        & \xmark  & \cmark& 32$\times$2  & SF-R101      & K700       & \cmark & 32.6   & 32.9  \\
STMixer-K                  & \xmark & \cmark & 16$\times$4  & ViT-B (from~\cite{vmae})        & K400       & \xmark & -               & 32.6           \\
STMixer-K                  & \xmark & \cmark & 16$\times$4  & ViT-B (from~\cite{VideoMAEv2})      & K710+K400       & \xmark & -               & 36.1           \\
STMixer-K        & \xmark  & \cmark & 16$\times$4  & ViT-L  (from~\cite{vmae})      & K700  & \xmark & -               & \textbf{39.5}  \\
\midrule
\multicolumn{8}{l}{\textbf{Compare to end-to-end methods}} & \\ \midrule            
AVA~\cite{ava}                   & \xmark  & \xmark& 20$\times$1  & I3D     & K400       & \xmark & 14.6            & -              \\
ACRN~\cite{acrn}                 & \xmark  & \xmark& 20$\times$1  & S3D-G        & K400       & \xmark & 17.4            & -              \\
STEP~\cite{step}                  & \xmark  & \xmark& 12$\times$1  & I3D      & K400       & \xmark & 18.6            & -              \\
VTr~\cite{vat}                   & \xmark  & \xmark& 64$\times$1  & I3D      & K400       & \xmark & 24.9            & -              \\
WOO~\cite{woo}                   & \xmark  & \xmark& 32$\times$2  & SF-R50       & K400       & \xmark & 25.2            & 25.4           \\
WOO~\cite{woo}                   & \xmark  & \xmark& 32$\times$2  & SF-R101-NL   & K600       & \xmark & 28.0            & 28.3           \\
TubeR~\cite{TubeR}                 & \xmark  & \cmark& 32$\times$2  & CSN-152      & IG-65M     & \xmark & 29.7            & 31.1           \\
TubeR~\cite{TubeR}                 & \xmark  & \cmark& 32$\times$2  & CSN-152      & IG-65M     & \cmark & 31.7            & 33.4           \\ \midrule
STMixer-K                  & \xmark & \cmark & 32$\times$2  & SF-R50       & K400       & \xmark & 27.2            & 27.8           \\
STMixer-K         & \xmark  & \cmark& 32$\times$2  & SF-R101-NL   & K600       & \xmark & 29.8   & 30.1  \\
STMixer-K                  & \xmark  & \cmark& 32$\times$2  & CSN-152      & IG-65M     & \xmark & 31.7            & 32.8           \\
STMixer-K        & \xmark  & \cmark& 32$\times$2  & CSN-152      & IG-65M     & \cmark & \textbf{34.4}   & 34.8  \\ \bottomrule
\end{tabular}
   }
    \label{tab:ava}
 \end{table*}
}

\newcommand{\TableAK}{
\begin{table}[!t]
    \centering
    \setlength{\tabcolsep}{3pt}
    \renewcommand{\arraystretch}{1.1}
    \caption{Comparisons with state-of-the-arts on AVA-Kinetics. \cmark of column ``Detector" denotes an extra human detector is required. * denotes test-time augmentation is used.}
    {
\begin{tabular}{@{}cccccc@{}}
\toprule
Method & Detector & Backbone & Input & Resolution & mAP \\ \midrule
SlowFast~\cite{slowfast,acarn} & \cmark & SF-R101 & $32\times 2$ & short side 256 & 33.0 \\
ACARN~\cite{acarn} & \cmark & SF-R101 & $32\times 2$ & short side 256 & 35.8 \\
*ACARN~\cite{acarn} & \cmark & SF-R101 & $32\times 2$ & short side 256 & 36.4 \\
Co-finetuning~\cite{coft} & \cmark & ViViT-B & $32\times 2$ & long side 512 & 36.2 \\
VTr~\cite{vat} & \xmark & I3D & $64\times 1$ & $400\times 400$ & 23.0 \\
STMixer-K & \xmark & ViT-B & $16\times 4$ & short side 256 & \textbf{37.5} \\ \bottomrule
\end{tabular}
   }
    \label{tab:ak}
 \end{table}
}

\newcommand{\TableTube}{
\begin{table*}[!t]
    \centering
    \setlength{\tabcolsep}{5.5pt}
    \renewcommand{\arraystretch}{1.1}
    \caption{Comparisons with state-of-the-arts on UCF101-24 and JHMDB51-21. \cmark of column ``Flow" denotes a 2-stream architecture using optical flow as extra input is adopted. Our STMixer-T surpasses former methods on all metrics by a clear margin.}
    {
 \begin{tabular}{@{}cccccccccccccc@{}}
\toprule
\multirow{3}{*}{Method} & \multirow{3}{*}{Backbone} & \multirow{3}{*}{Input} & \multirow{3}{*}{Flow} & \multicolumn{5}{c}{UCF101-24} & \multicolumn{5}{c}{JHMDB51-21} \\ \cmidrule(lr{0.5em}){5-9}\cmidrule(l{0.5em}r{0.5em}){10-14} 
 &  &  &  & \multirow{2}{*}{F-mAP@0.5} & \multicolumn{4}{c}{V-mAP} & \multirow{2}{*}{F-mAP@0.5} & \multicolumn{4}{c}{V-mAP} \\ \cmidrule(lr{0.5em}){6-9}\cmidrule(l{0.5em}r{0.5em}){11-14} 
 &  &  &  &  & @0.2 & @0.5 & @0.75 & 0.5:0.95 &  & @0.2 & @0.5 & @0.75 & 0.5:0.95 \\ \midrule
Saha \etal~\cite{saha} & VGG & $1\times 1$ & \cmark & - & 66.7 & 35.9 & 7.9 & 14.4 & - & 72.6 & 71.5 & 43.3 & 40.0 \\
Peng \etal~\cite{multitwo} & VGG & $1\times 1$ & \cmark & 39.9 & 42.3 & - & - & - & 58.5 & 74.3 & 73.1 & - & - \\
ROAD~\cite{online} & VGG & $1\times 1$ & \cmark & - & 73.5 & 46.3 & 15.0 & 20.4 & - & 73.8 & 72.0 & 44.5 & 41.6 \\
ACT~\cite{act}& VGG & $6 \times 1$ & \cmark & 69.5 & 76.5 & 49.2 & 19.7 & 23.4 & 65.7 & 74.2 & 73.7 & 52.1 & 44.8 \\
STEP~\cite{step} & VGG & $6 \times 1$ & \cmark & 75.0 & 76.6 & - & - & - & - & - & - & - & - \\
TACNet~\cite{tacnet} & VGG & $16 \times 1$ & \cmark & 72.1 & 77.5 & 52.9 & 21.8 & 24.1 & 65.5 & 74.1 & 73.4 & 52.5 & 44.8 \\
2in1~\cite{2in1} & VGG & $6\times 1$ & \cmark & - & 78.5 & 50.3 & 22.2 & 24.5 & - & - & 74.7 & 53.3 & 45.0 \\
ACRN~\cite{acrn}& S3D-G & $20\times 1$ & \cmark & - & - & - & - & - & 77.9 & - & 80.1 & - & - \\
AVA~\cite{ava} & I3D & $20\times 1$ & \cmark & 76.3 & - & 59.9 & - & - & 73.3 & - & 78.6 & - & - \\
CFAD~\cite{CFAD} & I3D & $32\times 1$ & \cmark & - & 81.6 & 64.6 & - & 26.7 & - & 86.8 & 85.3 & 63.8 & 53.0 \\ 
MOC~\cite{moc} & DLA34 & $7\times 1$ & \cmark & 78.0 & 82.8 & 53.8 & 29.6 & 28.3 & 70.8 & 77.3 & 77.2 & 71.7 & 59.1 \\
TubeR~\cite{TubeR} & I3D & $32\times 2$ & \cmark & 81.3 & 85.3 & 60.2 & - & 29.7 & - & 81.8 & 80.7 & - & - \\ \midrule
T-CNN~\cite{t-cnn} & C3D & $8\times 1$ & \xmark & 41.4 & 47.1 & - & - & - & 61.3 & 78.4 & 76.9 & - & - \\
CFAD~\cite{CFAD} & I3D & $32\times 1$ & \xmark & - & 79.4 & 62.7 & - & 25.5 & - & 84.8 & 83.7 & 62.4 & 51.8 \\
MOC\cite{moc} & DLA34 & $7\times 1$ & \xmark & 73.1 & 78.8 & 51.0 & 27.1 & 26.5 & - & - & - & - & - \\
TubeR~\cite{TubeR} & CSN-152 & $32\times 2$ & \xmark & 83.2 & 83.3 & 58.4 & - & 28.9 & - & 87.4 & 82.3 & - & - \\
STMixer-K & CSN-152 & $8\times 1$ & \xmark & 85.2  & 85.4 & 60.3 & 30.2 & 30.3 & 81.5 & 87.4 & 86.3 & 66.1 & 56.3 \\
STMixer-K & CSN-152 & $32\times 2$ & \xmark & 85.5 & 85.9 & 61.4 & 30.4 & 30.9 & \textbf{82.0} & 87.9 & 86.4 & 66.6 & 57.1 \\
STMixer-T & CSN-152 & $8\times 1$ & \xmark & \textbf{85.5} & \textbf{87.1} & \textbf{66.7} & \textbf{32.8} & \textbf{33.4} & 79.8 & \textbf{88.3} & \textbf{87.0} & \textbf{73.5} & \textbf{60.1} \\ \bottomrule
\end{tabular}
   }
    \label{tab:tube}
 \end{table*}
}

\newcommand{\TableMultiSports}{
\begin{table}[!t]
    \centering
    \setlength{\tabcolsep}{2.4pt}
    \renewcommand{\arraystretch}{1.1}
    \caption{Comparisons with state-of-the-arts on MultiSports. * denotes extra tracking model is used.}
    {
\begin{tabular}{@{}ccccccc@{}}
\toprule
 &  &  &  & \multicolumn{3}{c}{V-mAP} \\ \cmidrule(lr{0.5em}){5-7} 
\multirow{-2}{*}{Method} & \multirow{-2}{*}{Input} & \multirow{-2}{*}{Resolution} & \multirow{-2}{*}{F-mAP@0.5} & @0.2 & @0.5 & 0.1:0.9 \\ \midrule
ROAD~\cite{online,multisports} & $1\times 1$ & $300\times 300$ & 3.9 & 0.0 & 0.0 & - \\
YOWO~\cite{yowo,multisports} & $16\times 1$ & $224\times 224$ & 9.3 & 10.8 & 10.78 & - \\
MOC~\cite{moc,multisports} & $11\times 1$ & $288\times 288$ & 25.2 & 12.9 & 0.6 & - \\
SlowFast~\cite{slowfast,multisports} & $32\times 2$ & $256\times 455$ & 27.7 & 24.2 & 9.7 & - \\
HIT~\cite{hit} & $32\times 2$ & $256\times 455$ & 33.3 & 27.8 & 8.8 & - \\
PCCA~\cite{pcca} & $32\times 2$ & $256\times 455$ & 42.2 & 41.0 & 20.0 & 20.9 \\
TAAD~\cite{taad} & $32\times 2$ & $256\times 455$ & 49.6 & 54.1 & 31.3 & 28.9 \\
{\color[HTML]{C0C0C0} *TAAD~\cite{taad}} & {\color[HTML]{C0C0C0} $32\times 2$} & {\color[HTML]{C0C0C0} $256\times 455$} & {\color[HTML]{C0C0C0} 55.3} & {\color[HTML]{C0C0C0} 60.6} & {\color[HTML]{C0C0C0} 37.0} & {\color[HTML]{C0C0C0} 33.7} \\
STMixer-T & $10\times 1$ & $256\times 455$ & 50.4 & 57.7 & 35.2 & 32.0 \\
STMixer-T & $10\times 1$ & $288\times 512$ & \textbf{52.5} & \textbf{59.7} & \textbf{38.3} & \textbf{33.7} \\ \bottomrule
\end{tabular}
   }
    \label{tab:multisports}
 \end{table}
}